%% file: paper.tex
\documentclass[journal]{IEEEtran}

\usepackage{booktabs} 
\usepackage{color}
\usepackage{graphicx}
\usepackage{amsmath}
\usepackage{amssymb}
\usepackage{lettrine}
\usepackage[caption=false]{subfig}
\usepackage{url}

\newcommand{\etal}{\emph{et al.}}
 \newcommand{\edit}[1]{\textcolor{black}{#1}}

\begin{document}

\title{Deep Multi-Kernel Convolutional LSTM Networks and an Attention-Based Mechanism for Videos}  

\author{Sebastian Agethen, Winston H. Hsu\thanks{Sebastian Agethen and Winston H.~Hsu are affiliated with National Taiwan University.}\thanks{Email: d01944015@ntu.edu.tw; whsu@ntu.edu.tw}}

\maketitle


\begin{abstract}
Action recognition greatly benefits motion understanding in video analysis. Recurrent networks such as long short-term memory (LSTM) networks are a popular choice for motion-aware sequence learning tasks. Recently, a convolutional extension of LSTM was proposed, in which input-to-hidden and hidden-to-hidden transitions are modeled through convolution with a single kernel. This implies an unavoidable trade-off between effectiveness and efficiency. Herein, we propose a new enhancement to convolutional LSTM networks that supports accommodation of multiple convolutional kernels and layers. This resembles a Network-in-LSTM approach, which improves upon the aforementioned concern. In addition, we propose an attention-based mechanism that is specifically designed for our multi-kernel extension.
We evaluated our proposed extensions in a supervised classification setting on the UCF-101 and Sports-1M datasets, with the findings showing that our enhancements improve accuracy. We also undertook qualitative analysis to reveal the characteristics of our system and the convolutional LSTM baseline.\end{abstract}

\input{01.introduction/introduction.tex}
\input{02.related/related.tex}
\input{03.method/method.tex}
\input{04.evaluation/evaluation.tex}
\vspace{-0.1in}
\input{05.conclusion/conclusion.tex}
\vspace{-0.1in}
\section*{Acknowledgement}
\addcontentsline{toc}{section}{Acknowledgement}
This work was supported in part by the Ministry of Science and Technology, Taiwan, under Grant MOST 108-2634-F-002-004. We also benefit from the NVIDIA grants and the DGX-1 AI Supercomputer.

\bibliographystyle{IEEEtran}
\bibliography{paper}

\begin{IEEEbiography}[{\includegraphics[width=1in,height=1.25in,clip,keepaspectratio]{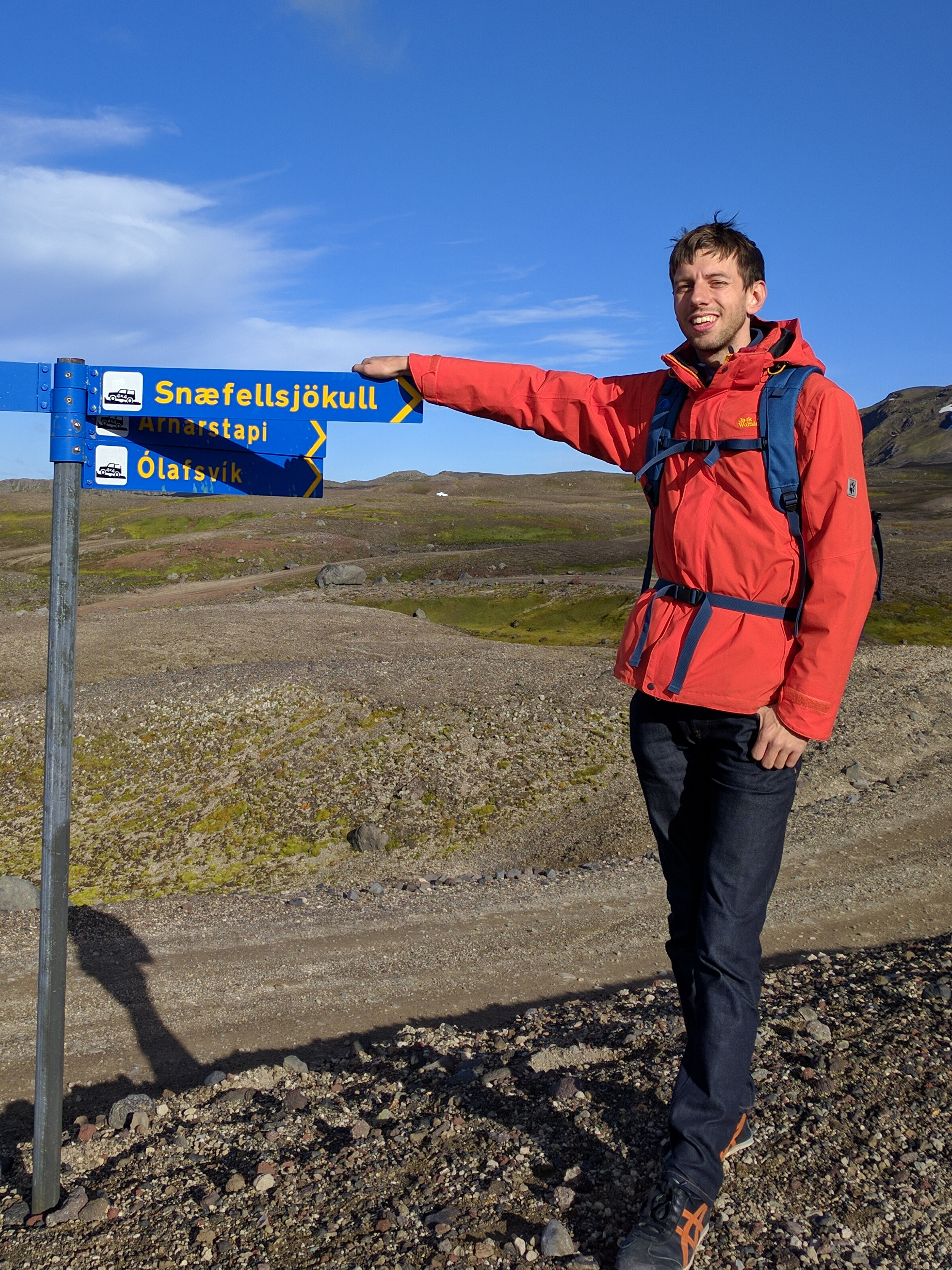}}]{Sebastian Agethen}
received his diploma in computer science in 2011 from RWTH Aachen University, Germany. He is currently working towards his Ph.D. degree at the Graduate Institute for Networking and Multimedia at National Taiwan University, Taiwan, which he expects to obtain in 2019. His research interests include deep learning, as well as human action recognition and prediction.
\end{IEEEbiography}
\vspace{-0.1in}
\begin{IEEEbiography}[{\includegraphics[width=1in,height=1.25in,clip,keepaspectratio]{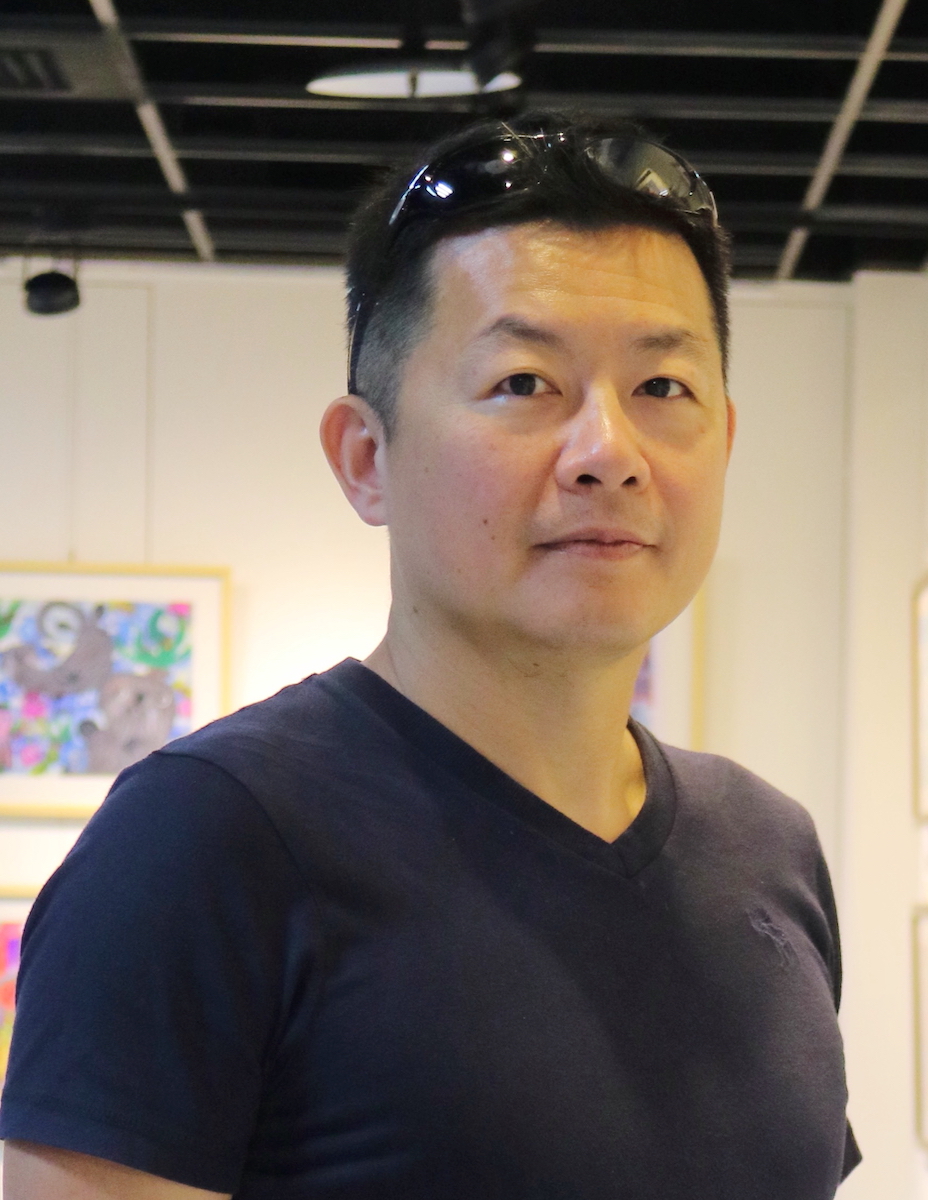}}]{Winston H. Hsu}
(S'03–M'07–SM'12) received the Ph.D. degree in electrical engineering from Columbia University, New York, NY, USA. He is keen to realizing advanced researches towards business deliverables via academia-industry collaborations and co-founding startups. Since 2007, he has been a Professor with the Graduate Institute of Networking and Multimedia and the Department of Computer Science and Information Engineering, National Taiwan University. His research interests include large-scale image/video retrieval/mining, visual recognition, and machine intelligence. Dr. Hsu served as the Associate Editor for the IEEE TRANSACTIONS ON MULTIMEDIA and on the Editorial Board for the IEEE MultiMedia Magazine.
\end{IEEEbiography}

\end{document}

%% file: 01.introduction/introduction.tex
\section{Introduction}
\label{sec:introduction}

\lettrine{\textbf{A}}CTION recognition is a challenging-yet-essential task in modern computer vision that is typically performed on video clips. Videos are now frequently encountered in our everyday lives on social media platforms such as Instagram, Facebook, and YouTube. The amount of video data is indeed vast; in 2015, 500 hours of videos were uploaded to YouTube every minute\footnote{According to a statement by YouTube CEO Susan Wojcicki at VidCon 2015.}. These quantities mean that automatic processing by machines is necessary, which is why action recognition is too. Many applications can benefit from action recognition; for example, autonomous driving, security and surveillance, and sports analysis.

Unlike static images, videos have an inherently spatiotemporal nature. The motion of subjects, such as persons, animals, or objects, carries significant information on the current action. By observing and exploiting motion, we can improve our understanding of an action over simple classification of static background features.

Particularly successful attempts at action recognition have been made using deep learning \cite{lit:beyondshortsnippets}\cite{lit:deepvideo}\cite{lit:twostream}\cite{lit:sequence2sequence}. Two approaches are common: deep convolutional networks (DCNs) have achieved impressive results, but they are  unaware of temporal dependencies (i.e., reordering frames has little effect), and therefore they perform poorly when processing motion information. A second choice is recurrent networks, among which long short-term memory \cite{lit:lstm} (LSTM) in particular has been used frequently in recent works. LSTM networks are sensitive to reordering of frames, and they can therefore be used for sequence learning.

\begin{figure}[t]
  \centering
  \def\svgwidth{\columnwidth}
  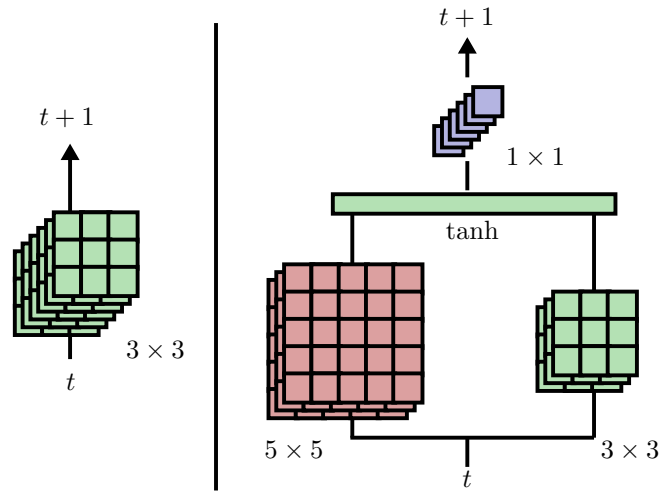
  \caption{Employing multiple kernels in ConvLSTM. \textbf{Left:} Traditional ConvLSTM with single kernel input-to-hidden and hidden-to-hidden transitions, here $3\times3$. \textbf{Right:} Exemplary multi-kernel configuration. We propose replacing the single kernel with a set of kernels. The total number of channels remains unchanged. In addition, a deep network may be used instead of a single layer.}
  \label{fig:multikernel}
\end{figure}

Several variations of LSTM exist. Most recently, convolutional LSTM cells \cite{lit:convlstm} (ConvLSTM) have been proposed, which add the benefits of convolution to an LSTM unit, namely \emph{spatial invariance\emph{;}\textit{\textit{}}} for example, given a two-dimensional image, a convolutional filter can correctly respond to an object no matter at which location it is placed in the image. This is not the case with \emph{fully connected} units, which were previously used in LSTM. In the following, we follow the notation in \cite{lit:convlstm} and refer to traditional LSTM as FC-LSTM, whereas we refer to convolutional LSTM as ConvLSTM. Our work is greatly based on ConvLSTM, and we discuss it in more detail in Section \ref{sec:method}.

In this paper, we reason that using a single kernel as in ConvLSTM is not optimal. First, whether a single optimal kernel size exists is unclear. Small kernels may not be able to build a causal connection between a phenomenon and a related previous one, as is the case when observing an actor's motion or with the simple example of a fast-moving object. Larger kernels, however, are very slow to compute and lack spatial invariance. We exhibit this in an example in Section~\ref{sec:method}. Second, we note a lack of depth, if we consider the transition at times $t \rightarrow t +1$: The convolutional operations performed can be viewed as a one-layer DCN. Additional depth and nonlinearities in particular are known to markedly improve learning.

To address these issues, we propose the following:
\begin{itemize}
  \item \textbf{Multi-kernel approach.} Replacement of the single kernel convolution with N output channels by a set of convolutional kernels of different dimensions (such that the number of output channels still sum up to N).
  \item \textbf{Additional depth.} Introduce additional depth for both \emph{input-to-hidden} and \emph{hidden-to-hidden} processing. For example, $1\times1$ convolutions may be used on the concatenated multi-kernel output, as seen in Fig.~\ref{fig:multikernel}. An alternative use is $1\times1$ bottlenecks ahead of large filters to reduce the computational burden.
  \item \textbf{Flow-based masking.} To support the multi-kernel approach, we propose generating kernel-dependent attention masks. One mask is generated for each convolutional kernel and applied pixelwise to the input before convolution.
\end{itemize}

To the best of our knowledge, such a Network-in-LSTM has not been used before. Attention-based models for different LSTM variants are known, but they differ from our approach, which is specifically tailored to support the multi-kernel approach.

The remainder of our paper is organized as follows: In Section~\ref{sec:related}, we discuss related work to our approach. We then present our method in detail in Section~\ref{sec:method}. In Section~\ref{sec:evaluation}, we comprehensively evaluate our approach, and we conclude our work in Section \ref{sec:conclusion}.

%% file: images/multikernel.pdf_tex
\begingroup%
  \makeatletter%
  \providecommand\color[2][]{%
    \errmessage{(Inkscape) Color is used for the text in Inkscape, but the package 'color.sty' is not loaded}%
    \renewcommand\color[2][]{}%
  }%
  \providecommand\transparent[1]{%
    \errmessage{(Inkscape) Transparency is used (non-zero) for the text in Inkscape, but the package 'transparent.sty' is not loaded}%
    \renewcommand\transparent[1]{}%
  }%
  \providecommand\rotatebox[2]{#2}%
  \ifx\svgwidth\undefined%
    \setlength{\unitlength}{273.4280804bp}%
    \ifx\svgscale\undefined%
      \relax%
    \else%
      \setlength{\unitlength}{\unitlength * \real{\svgscale}}%
    \fi%
  \else%
    \setlength{\unitlength}{\svgwidth}%
  \fi%
  \global\let\svgwidth\undefined%
  \global\let\svgscale\undefined%
  \makeatother%
  \begin{picture}(1,0.77149525)%
    \put(0,0){\includegraphics[width=\unitlength,page=1]{images/multikernel.pdf}}%
    \put(0.37784773,0.1){\color[rgb]{0,0,0}\makebox(0,0)[lt]{\begin{minipage}{0.42424319\unitlength}\raggedright $5\times5$\end{minipage}}}%
    \put(0.88251376,0.1){\color[rgb]{0,0,0}\makebox(0,0)[lt]{\begin{minipage}{0.59979209\unitlength}\raggedright $3\times3$\end{minipage}}}%
    \put(0,0){\includegraphics[width=\unitlength,page=2]{images/multikernel.pdf}}%
    \put(0.64,0.75){\color[rgb]{0,0,0}\makebox(0,0)[lt]{\begin{minipage}{0.49738856\unitlength}\raggedright $t+1$\end{minipage}}}%
    \put(0,0){\includegraphics[width=\unitlength,page=3]{images/multikernel.pdf}}%
    \put(0.675,0.055){\color[rgb]{0,0,0}\makebox(0,0)[lt]{\begin{minipage}{0.51472461\unitlength}\raggedright $t$\end{minipage}}}%
    \put(0,0){\includegraphics[width=\unitlength,page=4]{images/multikernel.pdf}}%
    \put(0.65,0.43){\color[rgb]{0,0,0}\makebox(0,0)[lt]{\begin{minipage}{0.78997007\unitlength}\raggedright $\tanh$\end{minipage}}}%
    \put(0.74067118,0.54){\color[rgb]{0,0,0}\makebox(0,0)[lt]{\begin{minipage}{0.35109781\unitlength}\raggedright $1\times1$\end{minipage}}}%
    \put(0,0){\includegraphics[width=\unitlength,page=5]{images/multikernel.pdf}}%
    \put(0.04,0.6){\color[rgb]{0,0,0}\makebox(0,0)[lt]{\begin{minipage}{0.45350134\unitlength}\raggedright $t+1$\end{minipage}}}%
    \put(0.08,0.2){\color[rgb]{0,0,0}\makebox(0,0)[lt]{\begin{minipage}{0.43887226\unitlength}\raggedright $t$\end{minipage}}}%
    \put(0.17,0.25){\color[rgb]{0,0,0}\makebox(0,0)[lt]{\begin{minipage}{0.43887228\unitlength}\raggedright $3\times3$\end{minipage}}}%
    \put(0,0){\includegraphics[width=\unitlength,page=6]{images/multikernel.pdf}}%
  \end{picture}%
\endgroup%

%% file: 02.related/related.tex
\section{Related Work}
\label{sec:related}

\subsubsection{Action recognition}
Action recognition has been a popular area of research for many years. To capture the information in videos, handcrafted global features such as the histogram of oriented gradients \cite{lit:hog} can be used. In the presence of noise, however, such features fail easily \cite{lit:hog}. To mitigate this issue, local features have increasingly been adopted \cite{lit:3dgradient}\cite{lit:bregonzio_spacetimeinterestpoints}\cite{lit:laptev_spacetimeinterestpoints}. Such local features focus on salient spatial regions in images; for example, a person, and can be temporally extended to some degree for videos. 

Most modern action recognition methods rely on deep features. In a breakthrough work, Karpathy \etal~\cite{lit:deepvideo} successfully investigated pooling strategies to fuse temporal information in DCNs. Further advances were made by Simonyan \etal~\cite{lit:twostream}, who showed that deep features extracted on optical flow can improve video classification. A notable insight of their work is the fact that convolutional networks based on optical flow features can be fine tuned from DCNs taught on RGB inputs, such as that used for image classification on the ImageNet dataset \cite{lit:imagenet}. The multistream architecture introduced in \cite{lit:twostream} has since been adopted by many works for action recognition \cite{lit:tmm-actionrecognition-2stream}\cite{lit:tmm-actionrecognition-3stream}.

The works by Donahue \etal~and Ng \etal~\cite{lit:longtermrecurrent}\cite{lit:beyondshortsnippets} successfully employed recurrent networks, specifically LSTM networks, to learn temporal sequences in action recognition videos.
Current approaches make extensive use of pretraining on large datasets. Carreira \etal~recently proposed inflated 3D convolutions \cite{lit:i3d}, and they demonstrated state-of-the-art performance by pretraining on large image and video datasets.

Several widely used datasets for action recognition are available. The most popular datasets have been UCF-101 \cite{lit:ucf101} and HMDB-51 \cite{lit:hmdb51}, which both contain thousands of videos and are therefore considered small scale. More recently, large-scale datasets such as Sports-1M \cite{lit:deepvideo}, Kinetics \cite{lit:kinetics}, and Moments in Time \cite{lit:momentsintime} have been published, each of which encompasses hundreds of thousands to millions of videos.

\subsubsection{Recurrent architectures and applications}
In this work, we investigate improvements to convolutional recurrent networks, in particular ConvLSTM \cite{lit:convlstm}. As traditional LSTM models suffer from a lack of spatial invariance by using fully connected operations, Shi \etal~addressed the issue by replacing the operations with convolutions and subsequently evaluating their system on a meteorological task. We detail the ConvLSTM equations in Section~\ref{sec:method}. 

Although they lack spatial invariance, fully connected LSTM-based models have also been widely used in fields other than action recognition. Sutskever \etal~\cite{lit:sequence2sequence} introduced the sequence-to-sequence framework to generate fixed-size representations from inputs of unconstrained sizes. The framework allows many applications; for example, the processing of natural language for translation purposes \cite{lit:googletranslation}\cite{lit:sequence2sequence}. Based on the same approach, several works have proposed solutions to tasks such as video-caption generation \cite{lit:s2vt}\cite{lit:lrcn}\cite{lit:videotranslation}.

Encoder--decoder structures, of which the sequence-to-sequence framework is one example, are frequently used for future prediction. Typically, an encoding LSTM reads in a sequence from the past or present, whereas a decoding LSTM produces the corresponding future. This can be used for low-dimensional tasks such as human pose prediction \cite{lit:theposeknows}. However, some work has also attempted to solve high-dimensional problems, such as the pixelwise reconstruction of future video frames; Srivastava \etal~\cite{lit:unsupervisedlstm} proposed an unsupervised learning method that is evaluated on the synthetic \emph{MovingMNIST} dataset, wherein synthetic video sequences are generated by assigning a handwritten digit (such as those in the MNIST dataset) a speed and an orientation. We use this intriguing dataset in our qualitative evaluation.
Future prediction tasks are also possible with the convolutional extension ConvLSTM; for example, future semantic segmentation \cite{lit:futuresemanticsegmentation} and weather prediction \cite{lit:deeprain}\cite{lit:convlstm}.

Attention-based mechanisms in (FC-)LSTM have proven useful for a variety of tasks. Gao~\etal~\cite{lit:tmm-attentionalrnn-captioning} used such a model for video captioning applications. Wang~\etal~\cite{lit:tmm-attentionalrnn-tracking} employed attention-based LSTM networks for a tracking task. Fan~\etal~\cite{lit:tmm-attentionalrnn-actionrecognition} conducted action recognition based on the human skeleton; their system architecture employed  LSTM networks to achieve that goal. Finally, Zhao~\etal~\cite{lit:tmm-attentionalrnn-objectclassification} used visual attention for fine-grained object classification.

Most recently, an attention system was attempted in connection with ConvLSTM in the work by Li~\emph{et al.}~\cite{lit:attentionconvlstm}. Notably, our use of attention masks differs from their work; although our approach is also motion based, we generate masks that are specialized for use with a specific convolutional kernel in a multi-kernel system.

\subsubsection{Inception}
Finally, our work bears some resemblance to the well-known GoogLeNet~\cite{lit:googlenet}, in particular the so-called \emph{Inception} module, as well as some resemblance with the network-in-network (NIN) approach \cite{lit:networkinnetwork}. An \emph{Inception} module is an ensemble of convolutional filters of different sizes (typically $1\times1$, $3\times3,$ and $5\times5$) in parallel. The module is designed purely with computational complexity (in terms of operations and parameters) in mind and applied in a DCN for \emph{static} image classification. By contrast, our work is designed under different considerations; we investigate the behavior of ConvLSTM for \emph{temporal} processes and argue that multiple kernels are necessary for correct recognition.

%% file: 03.method/method.tex
\section{Methodology}
\label{sec:method}

\subsection{Long Short-Term Memory}
In the following we briefly recapitulate both classical \emph{long short-term memory}~\cite{lit:lstm} (FC-LSTM) and the convolutional extension (ConvLSTM) presented in \cite{lit:convlstm}.

\subsubsection{Classic model}
Consider an input sequence $\mathbf{x}$ of length $T$, where $\mathbf{x}_t$ represents the $t$-th element. Such a sequence may for example be the RGB frames of a video clip or features extracted from a deep convolutional stack.
An LSTM unit is typically expressed as follows:
\begin{flalign}
  &\mathbf{i}_t  =  \sigma\left( \mathbf{W}_{xi}\mathbf{x}_t + \mathbf{W}_{hi}\mathbf{h}_{t-1} + \mathbf{W}_{ci} \circ \mathbf{c}_{t-1} + b_i \right)& \\
  &\mathbf{f}_t  =  \sigma\left( \mathbf{W}_{xf}\mathbf{x}_t + \mathbf{W}_{hf}\mathbf{h}_{t-1} + \mathbf{W}_{cf} \circ \mathbf{c}_{t-1} + b_f \right)& \\
  &\mathbf{o}_t  =  \sigma\left( \mathbf{W}_{xo}\mathbf{x}_t + \mathbf{W}_{ho}\mathbf{h}_{t-1} + \mathbf{W}_{co} \circ \mathbf{c}_{t-1} + b_o \right)& \\
  &\mathbf{c}_t  =  \mathbf{f}_t\circ\mathbf{c}_{t-1}+\mathbf{i}_t \circ \tanh\left(\mathbf{W}_{xc}\mathbf{x}_t + \mathbf{W}_{hc}\mathbf{h}_{t-1} + b_c\right)& \\
  &\mathbf{h}_t  =  \mathbf{o}_t \circ \tanh\left(\mathbf{c}_t\right)&
\end{flalign}
The tensor $\mathbf{c}_t$ is the so-called \emph{cell state}, whereas $\mathbf{h}_t$ is named the \emph{hidden state}, which acts as the output of the unit over time. The tensors $\mathbf{i}_t$, $\mathbf{f}_t,$ and $\mathbf{o}_t$ are the \emph{control gates}. We term the operations $\mathbf{W}_{x*} \cdot \mathbf{x}_t$ the \emph{input-to-hidden} transition and the operations $\mathbf{W}_{h*} \cdot \mathbf{h}_{t-1}$ the \emph{hidden-to-hidden} transition.
We also note that some studies and implementations may ignore the Hadamard terms $\mathbf{W}_{c*} \circ \mathbf{c}_{t-1}$. 

Backpropagation over time, which is used to train recurrent networks, suffers from the \emph{vanishing gradient}~\cite{lit:lstm} problem. The issue lies with the choice of activation functions; Sigmoid $\sigma(x)=\frac{1}{1+e^{-x}}$ and hyperbolic tangents $\tanh(x)=2\sigma(2x)-1$ have derivatives with upper bounds at 0.25 and 1, respectively, and repeated application therefore reduces the magnitude of a gradient.
Notably, the LSTM formulation above avoids this problem; given any output $\mathbf{h}_{t_i}$, we can find a path to any cell state $\mathbf{c}_{t_j}$ while passing only a single activation function such as $\tanh(\cdot)$ or $\sigma(\cdot)$.

Furthermore, we note the use of the dot product in the formulation. In effect, this resembles a fully connected layer as known from DCNs. Such fully connected layers are not spatially invariant \cite{lit:spatialinvariance}.

\subsubsection{Convolutional extension} To address the problem of spatial invariance, the authors of \cite{lit:convlstm} propose to replace the dot product by \emph{convolutional} operations. We refer to such a system in this work as ConvLSTM. The exact formulation is as follows:

\begin{flalign}
&\mathbf{i}_t  =  \sigma\left( \mathbf{W}_{xi} \ast \mathbf{x}_t + \mathbf{W}_{hi} \ast \mathbf{h}_{t-1} + \mathbf{W}_{ci} \circ \mathbf{c}_{t-1} + b_i \right)&\\
&\mathbf{f}_t  =  \sigma\left( \mathbf{W}_{xf} \ast \mathbf{x}_t + \mathbf{W}_{hf} \ast \mathbf{h}_{t-1} + \mathbf{W}_{cf} \circ \mathbf{c}_{t-1} + b_f \right)&\\
&\mathbf{o}_t  =  \sigma\left( \mathbf{W}_{xo} \ast \mathbf{x}_t + \mathbf{W}_{ho} \ast \mathbf{h}_{t-1} + \mathbf{W}_{co} \circ \mathbf{c}_{t-1} + b_o \right)&\\
&\mathbf{c}_t  =  \mathbf{f}_t \circ \mathbf{c}_{t-1} + \mathbf{i}_t \circ \tanh\left( \mathbf{W}_{xc} \ast \mathbf{x}_t + \mathbf{W}_{hc} \ast \mathbf{h}_{t-1} + b_c \right)&\\
&\mathbf{h}_t  =  \mathbf{o}_t \circ \tanh\left(\mathbf{c}_t\right)&
\end{flalign}

This modification ensures spatial invariance of the unit. The tensors $\mathbf{i},\mathbf{f},\mathbf{o},\mathbf{c,}$ and $\mathbf{h}$ are now convolutional maps and therefore higher dimensional.

\subsection{Problem of Kernel Sizes}

Each of the operations $\mathbf{W}_{x*} \ast \mathbf{x}_t$, $\mathbf{W}_{h*} \ast \mathbf{h}_{t-1}$ is associated with a single convolutional kernel. A problem that arises is the choice of optimal size; choosing a  kernel size that is too large results in the system degenerating into the original fully connected formulation. However, if the size is too small, the kernel will not be able to capture all information. 
In the original work on ConvLSTM, the authors \cite{lit:convlstm} found that a kernel size of $5\times5$ is optimal for the particular problem evaluated in their work, which was an unsupervised prediction problem such as weather prediction. 

In this paper, we argue that the exclusive use of kernels of one particular size is not optimal for ConvLSTM; instead we suggest using an array of kernels of different sizes, reminiscent of the \emph{Inception} module used with DCNs in \cite{lit:googlenet}.
Consider a video showing two objects, where the objects are moving at significantly different speeds. A small kernel is unable to link a fast-moving object during the transition from timestep $t$ to timestep $t+1$, simply because it has moved too far. However, always using large kernels is disadvantageous because they require \emph{more} parameters, are significantly \emph{slower} ($\frac{5\cdot5}{3\cdot3} \approx 2.78$), and \emph{degenerate} into fully connected layers at very large sizes, which lack spatial invariance. To solve this dilemma, we propose employing multiple kernels.

\begin{figure}[t]
  \centering
  \includegraphics[width=\columnwidth]{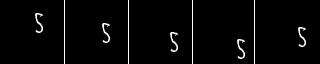}
  \caption{Sequence of a single handwritten digit moving over time (here with $T=5$). Note that the appearance of the digit remains unchanged at all times.}
  \label{fig:single_digit}
\end{figure}

\edit{
\subsubsection{Motion and kernel size}
\label{sec:motion} 
To support our hypothesis, we first visualize the problem of kernel size and velocity with the following simple experiment. In a variation of the \emph{MovingMNIST} dataset (which we discuss in more detail in Section~\ref{sec:attention_eval}), we construct a synthetic dataset of sequences of moving digits. We first pick a single digit from the MNIST dataset and subsequently assign it a velocity and a direction. The moving digit is then animated for T = 20 frames. The appearance of the digit remains unchanged at all times (i.e., we only translate it). On contact with the image boundaries, the moving digits are reflected. An example can be seen in Fig.~\ref{fig:single_digit}.}

\edit{
On this dataset, we now train a future predictor in the form of an encoder--decoder architecture. The encoding ConvLSTM reads in an input sequence of $T_{in}=10$ frames and produces a fixed-size state $(\mathbf{C},\mathbf{H})$. A second decoding ConvLSTM is initialized with this state and generates $T_{out}=10$ output features. A final $1\times1$ convolution layer reconstructs the predicted sequence. With each pixel taking on values in $[0,1]$, we can choose the binary sigmoid cross entropy (SCE) as our prediction loss:}
$$
L = \frac{-1}{N} \sum\limits_{n=1}^N \sum\limits_{t=1}^T p_n \log \hat{p}_n + (1-p_n) \log\left(1-\hat{p}_n\right)
$$
\edit{
To implement the encoder--decoder structure, we tried three different (traditional) ConvLSTM layers, each of them with a different kernel size (i.e., $3\times3$, $5\times5,$ and $7\times7$ kernels). We varied the average speed at which the digit was moving and repeated the experiment.}

\begin{figure}[t]
  \centering
  \includegraphics[width=1\columnwidth]{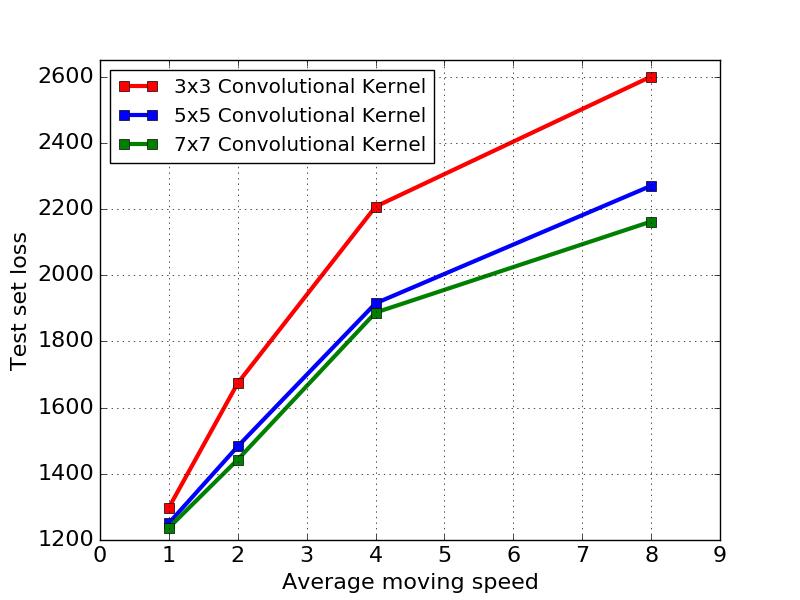}
  \caption{Test set (cross-entropy) loss for an unsupervised future prediction task with single moving digits (MovingMNIST). We vary the average moving speed of the digits but keep the appearance otherwise intact. Our observation is that the loss improves overproportionally for smaller convolutional kernels on slowly moving digits in comparison with their larger counterparts. Based on this, we propose that there is a correlation between moving speed and optimal kernel size.}
  \label{fig:correlation_result}
\end{figure}

\edit{
Fig.~\ref{fig:correlation_result} shows our results. The associated loss for smaller kernel sizes grew significantly faster than for larger kernel sizes. Concurrently, the computational costs were much higher for the larger kernels. We stress, however, that the digits in our experiment had static appearances, and in a real-world dataset motion may not necessarily be the only factor that determines the optimal kernel size.}

\begin{figure}
    \centering
    \subfloat[]{\includegraphics[width=.3\columnwidth]{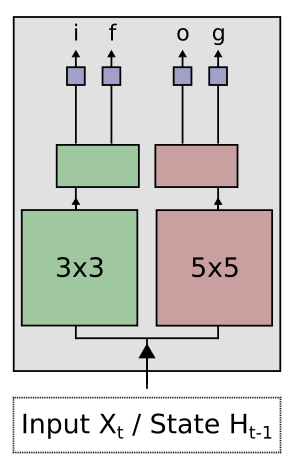}}
    \subfloat[]{\includegraphics[width=.3\columnwidth]{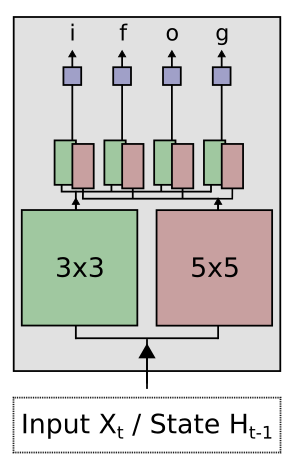}}
    \subfloat[]{\includegraphics[width=.3\columnwidth]{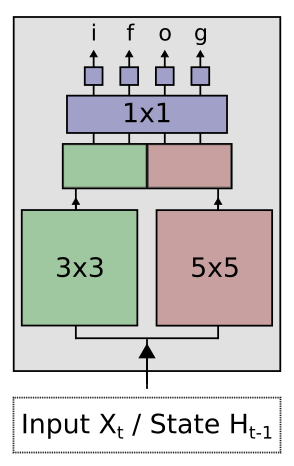}}
    
    \caption{\edit{Multi-kernel concatenation problem. Each gate activation $i,f,o,g$ should take a mixture of both kernels as input. (a) Using naive concatenation, each gate takes only one kernel output as input. (b) Splitting and interleaving before concatenation are one possible remedy for this. (c) An additional 1 x 1 convolution also avoids this problem and integrates both results.}}
    \label{fig:interleave}
\end{figure}

\edit{
\subsection{Concatenation of multiple kernels}
During implementation of a multi-kernel configuration, special care must be taken to correctly concatenate parallel kernels (see Fig.~\ref{fig:interleave}). From the ConvLSTM equations (6)--(9), we can see that the term inside the activation function (i.e., $\mathbf{W}_{x*}*\mathbf{x}_t + \mathbf{W}_{h*}*\mathbf{h}_{t-1} + b_*)$ differs only in the parameters $\mathbf{W},b$. A common implementation optimization is therefore to use two convolution operations $\mathbf{W}_x * \mathbf{x}_t$ and $\mathbf{W}_h * \mathbf{h}_{t-1}$ with $C' = 4\cdot C$ channels. The result is then split up into the four corresponding terms.
}

\edit{In our system, when using multiple kernels in parallel, we concatenate the individual results. At this point, the data order must be considered, as visualized in Fig.~\ref{fig:interleave} a). A naive concatenation will result in the gate activations $i,f,o,g$ to depend solely on one of the used kernels. However, each gate activation should take both kernel outputs as input. We propose two strategies to avoid this. First, as in Fig.~\ref{fig:interleave} b), we can interleave the results by first splitting each kernel's output into four and then concatenating in the correct order. Second, we can use a $1\times1$ convolution operation, which integrates the individual results (see Fig.~\ref{fig:interleave} c). Our results use the interleaving strategy in Fig.~\ref{fig:interleave} b) if no final $1\times1$ kernel is present.}

\edit{Our second strategy has the additional benefit that it introduces additional nonlinearity and depth, which is known to be beneficial \cite{lit:vgg} to deep learning. Convolutions with a $1\times1$ 
kernel can, however, also be used as so-called bottleneck layers: the layer is added to reduce the number of input channels before applying large convolutional kernels (e.g., a $5\times5$ kernel). We test bottleneck layers in our inception-like configuration in Section \ref{sec:i3d-based}.}

\subsection{Attention-based masking}
\label{sec:mask}
\begin{figure}[t]
  \centering
  \def\svgwidth{.8\columnwidth}
  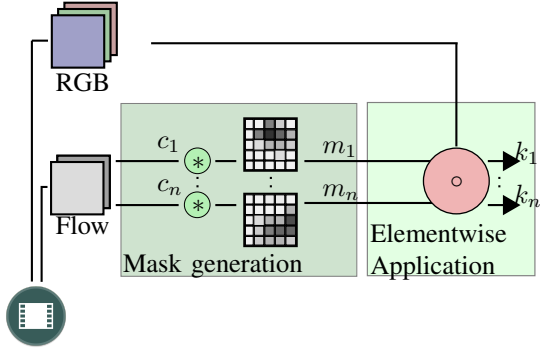
  \caption{Multi-kernel attention scheme. For each kernel using attention, we generate a mask $m_i$ by convolution $c_i$ over the flow features. Subsequently, for each ConvLSTM kernel $k_i$, the corresponding mask is applied to the RGB input features by element-wise multiplication.}
  \label{fig:attentionmask}
\end{figure}

In our multi-kernel extension, any kernel learns on all regions of an image, even if it is not optimally suited for this image region. Ideally, our system should avoid this. In this work, we aim to enforce large kernels to concentrate on faster objects and smaller kernels to concentrate on slower objects. Note that the background scene and static objects may also contribute to the learning process, and we consequently suggest to apply attention masks only on a subset of convolutional kernels.

To determine the utility of a kernel on a particular image region, we generate attention masks from optical flow features. The magnitude of the optical flow determines the distance a particular pixel has moved on the $x$- or $y$-axis, and therefore represents speed. To generate each mask, we employ a DCN on the optical flow features. A non-linearity, such as the sigmoid, is applied subsequently.
This mask can then be applied by element-wise multiplication to all input features $\mathbf{x}_t$. 
Our goal is that through backpropagation of error, each mask specializes onto the corresponding convolutional kernel. Fig.~\ref{fig:attentionmask} shows this process. 

The above masking process affects input-to-hidden convolutions (i.e., parameters $\mathbf{W}_{x*}$), and we reserve a hidden-to-hidden extension ($\mathbf{W}_{h*}$) for future work.

%% file: images/attention_mask.pdf_tex
\begingroup%
  \makeatletter%
  \providecommand\color[2][]{%
    \errmessage{(Inkscape) Color is used for the text in Inkscape, but the package 'color.sty' is not loaded}%
    \renewcommand\color[2][]{}%
  }%
  \providecommand\transparent[1]{%
    \errmessage{(Inkscape) Transparency is used (non-zero) for the text in Inkscape, but the package 'transparent.sty' is not loaded}%
    \renewcommand\transparent[1]{}%
  }%
  \providecommand\rotatebox[2]{#2}%
  \ifx\svgwidth\undefined%
    \setlength{\unitlength}{406.8bp}%
    \ifx\svgscale\undefined%
      \relax%
    \else%
      \setlength{\unitlength}{\unitlength * \real{\svgscale}}%
    \fi%
  \else%
    \setlength{\unitlength}{\svgwidth}%
  \fi%
  \global\let\svgwidth\undefined%
  \global\let\svgscale\undefined%
  \makeatother%
  \begin{picture}(1,0.65334186)%
    \put(0,0){\includegraphics[width=\unitlength,page=1]{images/attention_mask.pdf}}%
    \put(0.1,0.25){\color[rgb]{0,0,0}\makebox(0,0)[lt]{\begin{minipage}{0.29498525\unitlength}\raggedright Flow\end{minipage}}}%
    \put(0.1,0.52){\color[rgb]{0,0,0}\makebox(0,0)[lt]{\begin{minipage}{0.45231072\unitlength}\raggedright RGB\end{minipage}}}%
    \put(0,0){\includegraphics[width=\unitlength,page=2]{images/attention_mask.pdf}}%
    \put(0.44740117,1.58953786){\color[rgb]{0,0,0}\makebox(0,0)[lt]{\begin{minipage}{0.18682399\unitlength}\raggedright \end{minipage}}}%
    \put(0,0){\includegraphics[width=\unitlength,page=3]{images/attention_mask.pdf}}%
    \put(0.355,0.3625){\color[rgb]{0,0,0}\makebox(0,0)[lt]{\begin{minipage}{0.29498524\unitlength}\raggedright $\ast$\end{minipage}}}%
    \put(0,0){\includegraphics[width=\unitlength,page=4]{images/attention_mask.pdf}}%
    \put(0.355,0.281){\color[rgb]{0,0,0}\makebox(0,0)[lt]{\begin{minipage}{0.29498524\unitlength}\raggedright $\ast$\end{minipage}}}%
    \put(0,0){\includegraphics[width=\unitlength,page=5]{images/attention_mask.pdf}}%
    \put(0.22615536,0.18){\color[rgb]{0,0,0}\makebox(0,0)[lt]{\begin{minipage}{0.84562434\unitlength}\raggedright Mask generation\end{minipage}}}%
    \put(0.29,0.4){\color[rgb]{0,0,0}\makebox(0,0)[lt]{\begin{minipage}{\unitlength}$c_1$\end{minipage}}}%
    \put(0.29,0.32){\color[rgb]{0,0,0}\makebox(0,0)[lt]{\begin{minipage}{\unitlength}$c_n$\end{minipage}}}%
    \put(0.6,0.39){\color[rgb]{0,0,0}\makebox(0,0)[lt]{\begin{minipage}{\unitlength}$m_1$\end{minipage}}}%
    \put(0.6,0.31){\color[rgb]{0,0,0}\makebox(0,0)[lt]{\begin{minipage}{\unitlength}$m_n$\end{minipage}}}%
    \put(0.96,0.39){\color[rgb]{0,0,0}\makebox(0,0)[lt]{\begin{minipage}{\unitlength}$k_1$\end{minipage}}}%
    \put(0.96,0.31){\color[rgb]{0,0,0}\makebox(0,0)[lt]{\begin{minipage}{\unitlength}$k_n$\end{minipage}}}%
    \put(0.6902655,0.235){\color[rgb]{0,0,0}\makebox(0,0)[lt]{\begin{minipage}{0.4818093\unitlength}\raggedright Elementwise\\ Application\end{minipage}}}%
    \put(0,0){\includegraphics[width=\unitlength,page=6]{images/attention_mask.pdf}}%
    \put(0.84,0.325){\color[rgb]{0,0,0}\makebox(0,0)[lt]{\begin{minipage}{0.18682405\unitlength}\raggedright $\circ$\end{minipage}}}%
    
  \end{picture}%
\endgroup%

%% file: 04.evaluation/evaluation.tex
\section{Evaluation}
\label{sec:evaluation}
In the following, we present our experimental results. We consider both quantitative and qualitative aspects. First, we seek a direct comparison with the ConvLSTM baseline, and we show on the example of video classification that our proposed method improves classification accuracy. Following this, we attempt to gain a better understanding of our modifications in the qualitative analysis.
\vspace{-0.1in}
\subsection{Quantitative results}
The work in \cite{lit:convlstm} was conducted with weather forecasting in mind. Our proposed changes, however, benefit applications in which visible objects move at different speeds. Hence, we chose action recognition as the application to quantitatively evaluate our system. We begin with a description of the two popular action recognition datasets used in this work: UCF-101 \cite{lit:ucf101} and Sports-1M \cite{lit:deepvideo}.

\begin{figure}
  \centering
  \def\svgwidth{.8\columnwidth}
  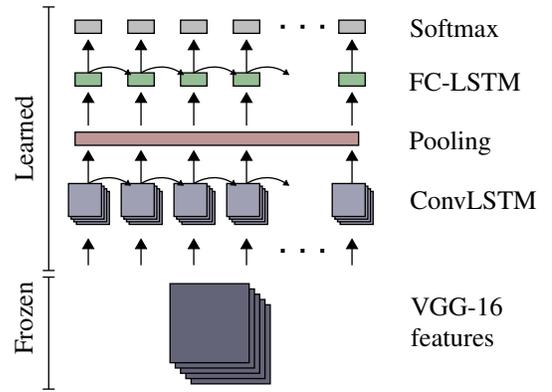
  \caption{Model used for all VGG-based experiments on Sports-1M and UCF-101.}
  \label{fig:recurrentstack}
\end{figure}

\subsubsection{Datasets}
The \emph{UCF-101} action recognition dataset contains 13,320 videos, each showing a single human action. The average length of the videos is 180 frames (i.e., just a few seconds), and there are 101 categories, which implies a small number of videos per class, slightly more than 100 on average. \edit{This causes models trained solely on UCF-101 to overfit easily.} 
The dataset offers three different splits into training and test data; we used split one.

The second dataset, the\emph{Sports-1M dataset}, is a collection of over one million videos on YouTube in 487 classes, which all relate to sports. The videos have unconstrained lengths, ranging from a few seconds to hours.
The size of the dataset inherently causes difficulties; required storage space is massive even on the lowest resolution (approximately~7 TB), and processing one million videos in each epoch leads to long computation times. For this study, we therefore decided to run our experiments on a subset by randomly sampling 20 classes\footnote{\emph{rafting, skittles, test cricket, shidokan, pitch and putt, dirt track racing, freestyle skiing, street football, sprint, motorcycle speedway, trial, dressage, surf fishing, juggling club, soft tennis, sailing, road racing, jetsprint, gatka,} and \emph{enduro.}}.

\begin{figure*}
  \includegraphics[width=\linewidth]{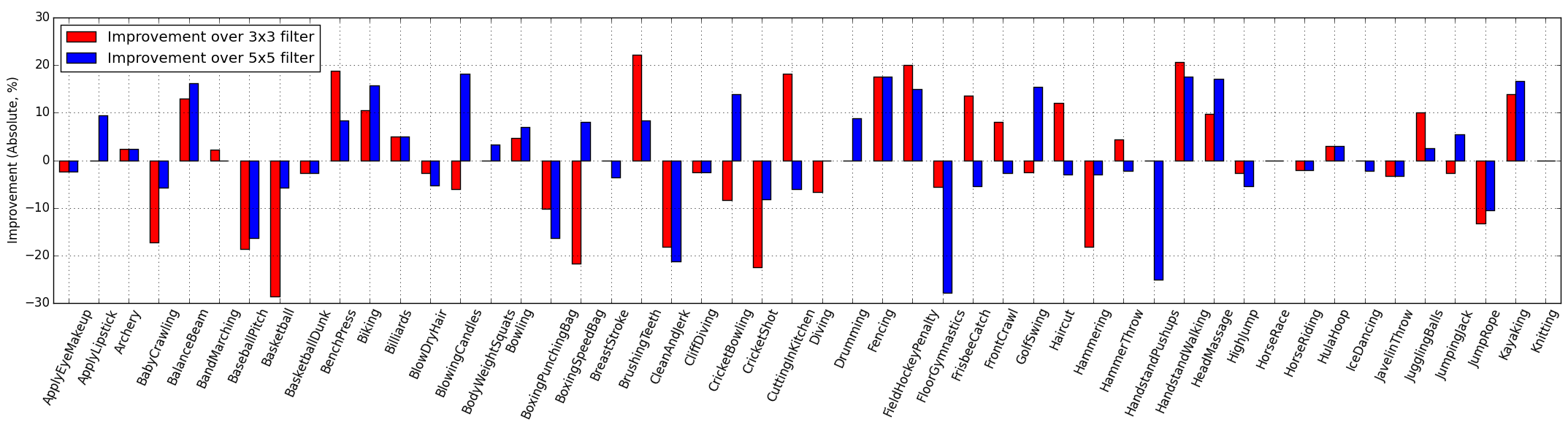}
  \includegraphics[width=\linewidth]{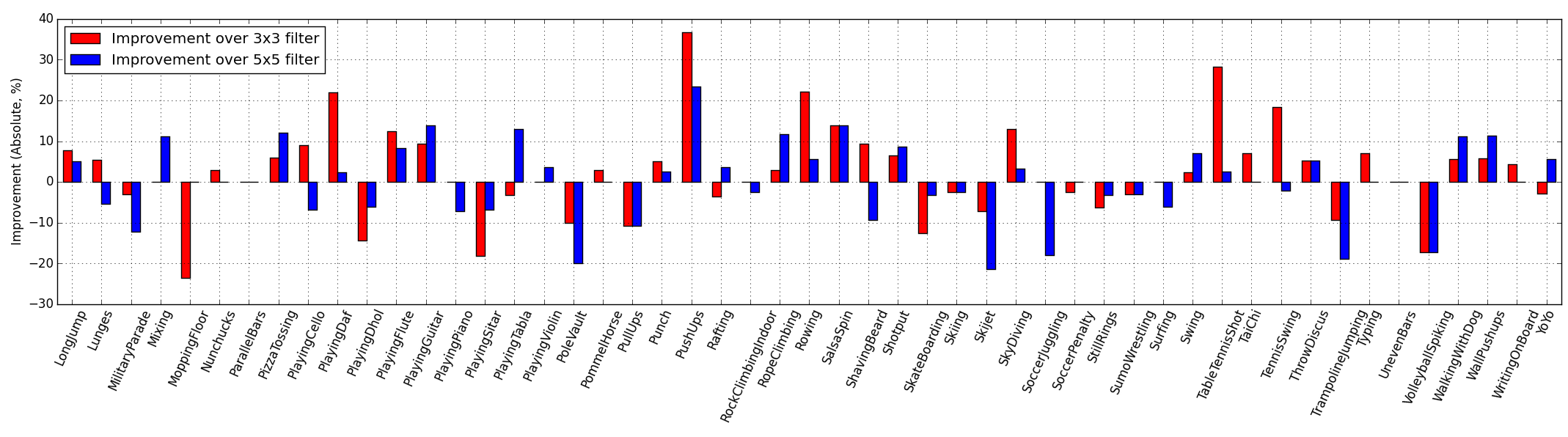}
  \caption{Classwise comparison of our multi-kernel system with two ConvLSTM baselines on the UCF-101 test set (split 1). Notably,  classes 27 and 71 (\texttt{Fencing}, \texttt{PushUps}) strongly outperform both baselines.}
  \label{fig:classwise}
\end{figure*}

\subsubsection{Experimental setup}

\edit{Two difficulties arose during our experiments. First, unrolling a convolutional LSTM over time requires large amounts of GPU memory. Second, training on a small-scale dataset such as UCF-101 will lead to severe overfitting when training from scratch. To mitigate these two problems, we therefore decided to initialize the weights of the convolutional stack with pretrained weights and subsequently freeze them. Here, we describe our setup for the two convolutional base network variants.}

\edit{
The first set of experiments used a VGG-16~\cite{lit:vgg} convolutional stack. We used ImageNet-pretrained weights and extracted features of dimensions $14\times14\times512$ at layer \texttt{conv5\_3}. Using the larger spatial extent in this layer exposes the differences between smaller and larger kernels more clearly. A ConvLSTM unit then processed the features. We established both $3\times3$ and $5\times5$ kernel baselines with $C=512$ channels and then compared with a mixture of both kernels ($C_{3\times3}=256, C_{5\times5}=256$) while keeping the total number of channels fixed. The ConvLSTM layer was followed by pooling, an FC-LSTM, and a Softmax classifier.}

\edit{
Our second set of experiments employed a recent state-of-the-art action recognition network named I3D~\cite{lit:i3d}. Here, we initialized with Kinetics-pretrained weights and extracted RGB features of dimensions $7\times7\times1024$ at layer \texttt{incept5b}. 
Accuracy was averaged over all frames in both sets of experiments.}

\subsubsection{Results of VGG-based configurations}

\begin{table}
  \caption{Supervised classification results of UCF-101 on top of a VGG~\cite{lit:vgg} architecture. Our multi-kernel configuration clearly outperforms both baseline variants. }
  
  \label{tbl:ucf101}
  \centering
  \begin{tabular}{|l|c|}
    \hline
    \textbf{Configuration}                             & \textbf{Top-1 Accuracy} \\ \hline
    Baseline $3\times3\times512$                       & 71.27\% \\ \hline
    Baseline $5\times5\times512$                       & 72.20\% \\ \hline \hline
    Simple Multi-kernel                                & 73.18\% \\ 
    $C_3 = C_5 = 256$ \textbf{(Ours)}                  & \\ \hline
    Simple Multi-kernel (with stacked $1\times1$)      & \textbf{74.09\%} \\ 
    $C_3 = C_5 = 256$ \textbf{(Ours)}                  & \\ \hline
  \end{tabular}
\end{table}

UCF-101 contains very short videos, and we therefore sampled them at 15 fps. From a set of 140 extracted frames, we chose $T=50$ (i.e., frames 25 to 75) frames as input. Our results for this dataset can be found in Table~\ref{tbl:ucf101}.

\begin{table}
  \centering
  \caption{Time per iteration of ConvLSTM layer with $C_{VGG}=512$ input channels, batch size of 10, and T = 20 frames per video. The numbers are averaged over 100 iterations.}
  \label{tbl:complexity_timings}
  \begin{tabular}{|l|c|c|}
    \hline
    \textbf{Configuration}  & \textbf{Training}  & \textbf{Testing} \\ \hline
    $3\times3\times256$     & 3.424s    & 2.821s             \\ \hline
    $5\times5\times256$     & 4.430s    & 3.290s             \\ \hline
    \hline
    Simple Multi-kernel     & 4.007s    & 3.008s  \\
    $C_3 = C_5 = 128$ \textbf{(Ours)}      & & \\ \hline
    Simple Multi-kernel (with stacked $1\times1$)  & 4.176s     & 3.072s \\ 
    $C_3 = C_5 = 128$ \textbf{(Ours)}      &           &       \\ \hline
  \end{tabular}
\end{table}

\begin{table}
  \centering
  \caption{Number of parameters required for different ConvLSTM configurations. We assume $C_{VGG}=512$ input channels.}
  \label{tbl:complexity_parameters}
  \begin{tabular}{|l|c|c|}
    \hline
    \textbf{Configuration} & \textbf{Weights} & \textbf{Biases} \\ \hline
    $3\times3\times256$ & 4.7 M   & 1024 \\ \hline
    $5\times5\times256$ & 13.1 M  & 1024 \\ \hline
    $7\times7\times256$ & 25.7 M  & 1024 \\ \hline \hline
    Simple Multi-kernel & 8.9 M & 1024 \\ 
    $C_3 = C_5 = 128$ \textbf{(Ours)}      & & \\ \hline
    Simple Multi-kernel (with stacked $1\times1$) & \textbf{9.1 M} & \textbf{1024} \\
    $C_3 = C_5 = 128$ \textbf{(Ours)}      & & \\ \hline
  \end{tabular}
\end{table}

Videos in Sports1M-20 are typically longer, and we therefore sampled frames at 1 fps. We sampled a set of 140 frames and picked $T=30$ frames (i.e., frames 70 to 100) as input for our system. In cases where the videos were not sufficiently long, those frames were sampled modulus their length. Our results are presented in Table \ref{tbl:sports1m20}.

\begin{table}
  \caption{Supervised classification results of Sports1M-20 on top of a VGG~\cite{lit:vgg} architecture. Both baseline and the proposed method each have a total of $C=512$ output feature maps.}
  \label{tbl:sports1m20} 
  \centering
  \begin{tabular}{|l|c|}
    \hline
    \textbf{Configuration}           & \textbf{Top-1 Accuracy} \\ \hline
    Baseline $3\times3\times512$ & 80.67\% \\ \hline
    Baseline $5\times5\times512$ & 81.09\% \\ \hline \hline
    Simple Multi-kernel               & \textbf{81.34\%} \\ 
    $C_3 = C_5 = 256$ \textbf{(Ours)} & \\ \hline

  \end{tabular}
\end{table}

\begin{figure}
  \centering
  \includegraphics[width=.75\columnwidth]{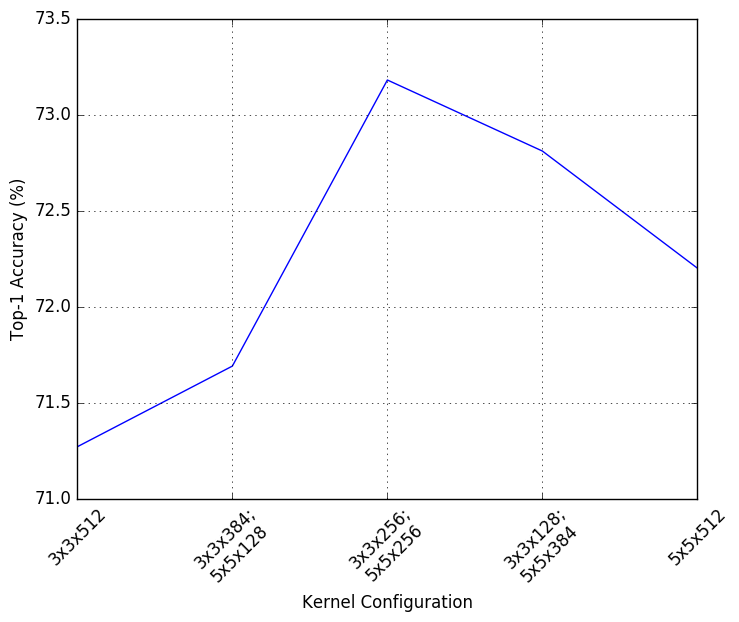}
  \caption{Top-1 accuracy on the UCF-101 dataset with varying proportions of $3\times3$ and $5\times5$ convolutional kernels.}
  \label{fig:kernelconfigurations}
\end{figure}

Our experiment evaluated the performance of the proposed multi-kernel system and compared it against the ConvLSTM baselines, which use a single convolutional kernel only. Our approach outperformed both the $3\times3$ and $5\times5$ ConvLSTM baselines, with improvements of 2.82\% and 1.89\% on UCF-101, respectively. The improvement was more marginal on Sports-1M, where we gained 0.67\%\ and 0.25\%, respectively. In both cases, the larger $5\times5$ baseline performed better than the $3\times3$ baseline, as already observed in Fig. \ref{fig:single_digit}. Furthermore, we attempted several configurations of $3\times3$ and $5\times5$ kernels for our multiple kernel system on UCF-101, and the results are shown in Fig.~\ref{fig:kernelconfigurations}.

\edit{Both baseline and our extension performed well below state-of-the-art levels on UCF-101. Modern state-of-the-art networks utilize extensive pretraining (e.g., on Kinetics as in \cite{lit:i3d}) and batch normalization \cite{lit:batchnorm} or dropout \cite{lit:dropout}. Our proposed extension can be combined with such modern networks easily. Here, we report the results of experiments with an I3D-based setup and show that state-of-the-art can be beaten on a subset of UCF-101.}

\subsubsection{Results of I3D-based configurations}
\label{sec:i3d-based}
\begin{figure}
    \centering
    \subfloat[]{\includegraphics[height=1.5in]{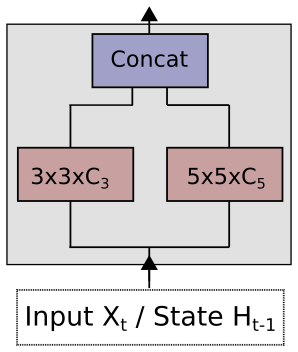}}
    \subfloat[]{\includegraphics[height=1.5in]{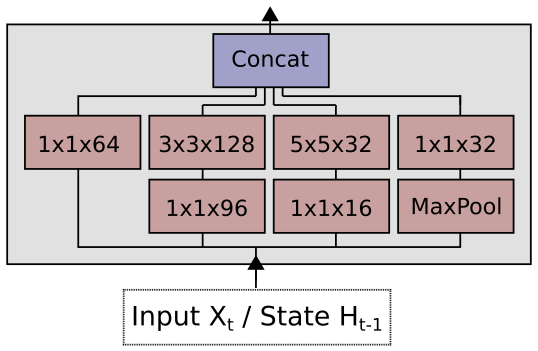}}        
    \caption{Configurations used in our experiments: (a) simple multi-kernel and (b) Inception-like multi-kernel.}
    \label{fig:inception-like}
\end{figure}

\edit{We first show that standard I3D can be outperformed on the RGB stream on a subset\footnote{ApplyEyeMakeup, BlowDryHair, BoxingPunchingBag, FloorGymnastics, HandstandPushups, HorseRace, JumpRope, Lunges, MilitaryParade, PlayingCello, PlayingPiano, PlayingTabla, PushUps, Rafting, Shotput, Skiing, WalkingWithDog} of classes of UCF-101. According to \cite{lit:i3d}, I3D greatly profits from high temporal resolution, and consequently, we sampled frames at 24 fps, selecting clips of 64 frames. The pooling and FC-LSTM used in the VGG configuration were removed and replaced with a $1\times1$ convolution with batch normalization. 
We tested a configuration similar to the Inception layer in \cite{lit:googlenet} (see Fig.~\ref{fig:inception-like} b). We used $C=256$ in all cases. Because batch normalization can combat overfitting, we supplied all convolutional kernels with such normalization. Our results for a 17-classes subset of UCF-101 are presented in Table~\ref{tab:i3d}; both of our configurations outperformed I3D, which is state-of-the-art, by over 0.84\%. }

\begin{table}[h]
    \centering
    \edit{
    \caption{State-of-the-art experiment on a 17-class subset of UCF-101. Our Inception-like multi-kernel outperforms standard I3D by 0.84\% as well as both ConvLSTM baselines on the RGB stream. Further improvement can principally be achieved by pretraining the ConvLSTM layer, which is outside the scope of this study. For nomenclature, please refer to Fig.~\ref{fig:inception-like}.}
    \begin{tabular}{|ll|r|}
    \hline
        \textbf{Configuration} &       & \textbf{Top-1 accuracy} \\ \hline
        I3D                 & Baseline &  96.62\% \\  \hline
        ConvLSTM $3\times3$ & Baseline &  96.46\% \\ \hline
        ConvLSTM $5\times5$ & Baseline &  96.62\% \\ \hline \hline
        Inception multi-kernel & \textbf{Ours} & \textbf{97.46\%} \\ \hline
    \end{tabular}
    \label{tab:i3d}
    }
\end{table}

\begin{table}[]
    \centering
  \edit{
    \caption{Classification results on UCF-101, choosing I3D as the base network: \textbf{End-to-end} training and ablative study for our \textbf{flow-based attention} scheme.}
    \begin{tabular}{|ll|r|}
    \hline
        \textbf{Configuration}          &    & \textbf{Top-1 accuracy} \\ \hline
        ConvLSTM $3\times 3$            & Baseline       & 86.33 \% \\ \hline
        ConvLSTM $5\times 5$            & Baseline       & 86.92 \% \\ \hline \hline
        Simple multi-kernel             & \textbf{Ours}   & 87.21\% \\ \hline
        Simple multi-kernel             & \textbf{Ours}   & \textbf{87.39\%} \\
        (\emph{with} flow-based attention) &  &         \\ \hline \hline
        Inception multi-kernel          & \textbf{Ours}& 88.40\% \\ \hline
        Inception multi-kernel          & \textbf{Ours}& \textbf{90.09\%} \\ 
        (trained end-to-end)            &        & \\ \hline
    \end{tabular}
    \label{tab:additional}
  }
\end{table}

\edit{We also evaluated our proposed optical flow-based attention. Using an I3D base network and the simple multi-kernel configuration $C_{3\times3}=128, C_{5\times5}=128$, we generated attention masks and applied them to the input features of the ConvLSTM. The result in Table~\ref{tab:additional} shows an improvement of 0.18\%. We used a single convolutional layer to generate the attention masks. In our future work, we will consider deeper architectures to improve flow masks further.}

\edit{Finally, we trained an Inception multi-kernel in an end-to-end fashion by simultaneously fine tuning the I3D base network. The Inception multi-kernel was an excellent choice for this experiment because $1\times1$ bottleneck convolutions help reduce the memory footprint significantly. Table~\ref{tab:additional} shows that training in an end-to-end fashion improved accuracy by a further 1.69\%.}

\subsection{Analysis}

Because  our proposed extensions may perform better on certain data, we also provide a breakdown of classification accuracy by class for UCF-101 (with the VGG-based setup). Our system improved classification if both the blue and red bars in Fig.~\ref{fig:classwise} show positive values. Considerable improvements can be seen in certain categories, such as \texttt{Fencing} or \texttt{Pushups}. In other cases, a class did not benefit from multiple kernels. We suggest that this occurs if motion velocity is fairly constant throughout the instances of a class. Such an example is \texttt{MoppingFloor}, where a pure $3\times3$ ConvLSTM exhibits better performance.

\begin{figure*}
  \includegraphics[width=\linewidth]{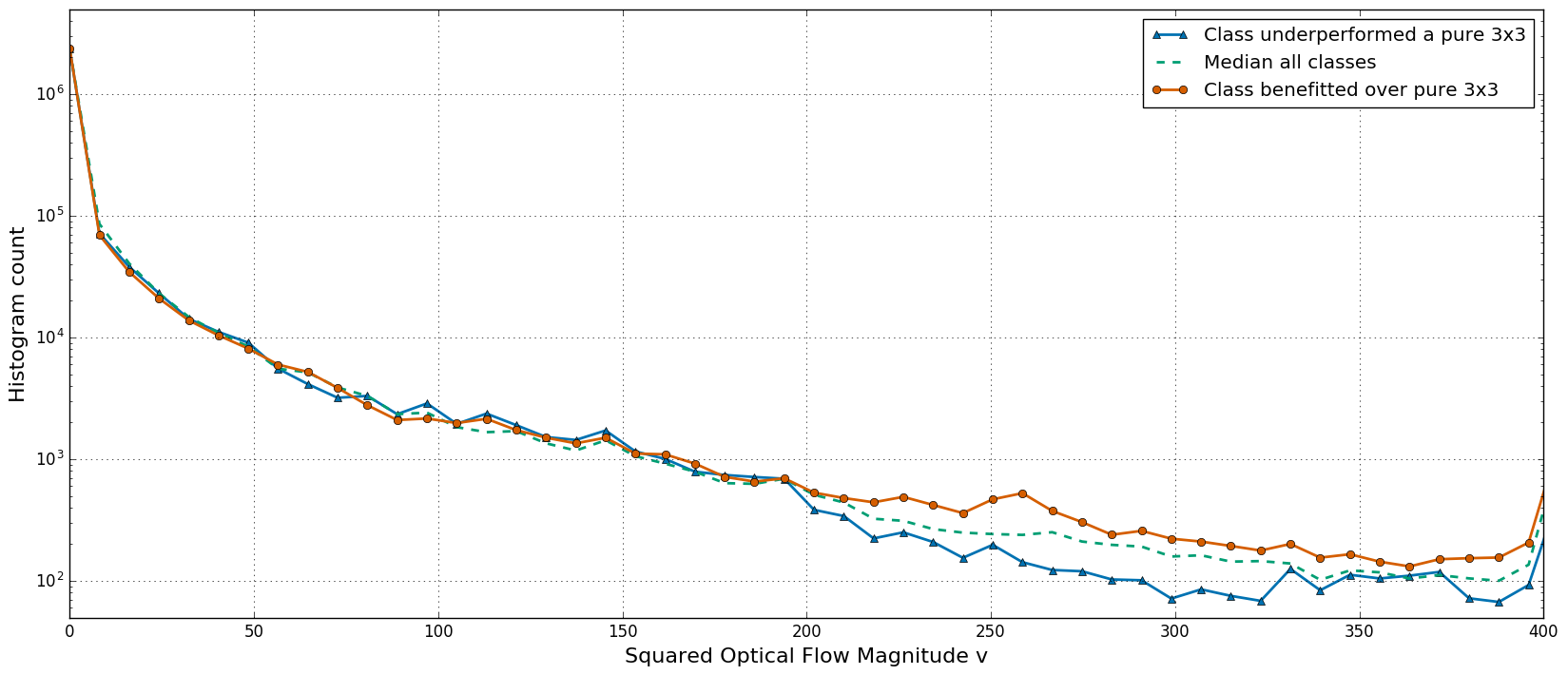}
  \caption[Histogram 3x3]{Histogram of squared optical flow velocities $v=\left(x^2 + y^2\right)$ for two sets of classes. The blue line shows the distribution for classes that performed significantly better in the ConvLSTM $3\times3$ baseline, whereas the red line shows the counts for classes in which our approach outperformed the $3\times3$ baseline. For velocities $200 \leq v \leq 400,$ the blue line is significantly below the median of all classes, while the red line is above the median. In other words, our extensions work best when many velocities between 200 and 400 are present. Note that the peak near $v=400$ is because we clipped all $x, y$ values with magnitudes above 20, and we focus on $0 \leq v \leq 400$ here (the maximum value of $v$ is 800). Furthermore. the histogram counts are normalized by number of videos per class.} 
  \label{fig:histogram_velo_3}
\end{figure*}

\begin{figure*}
  \includegraphics[width=\linewidth]{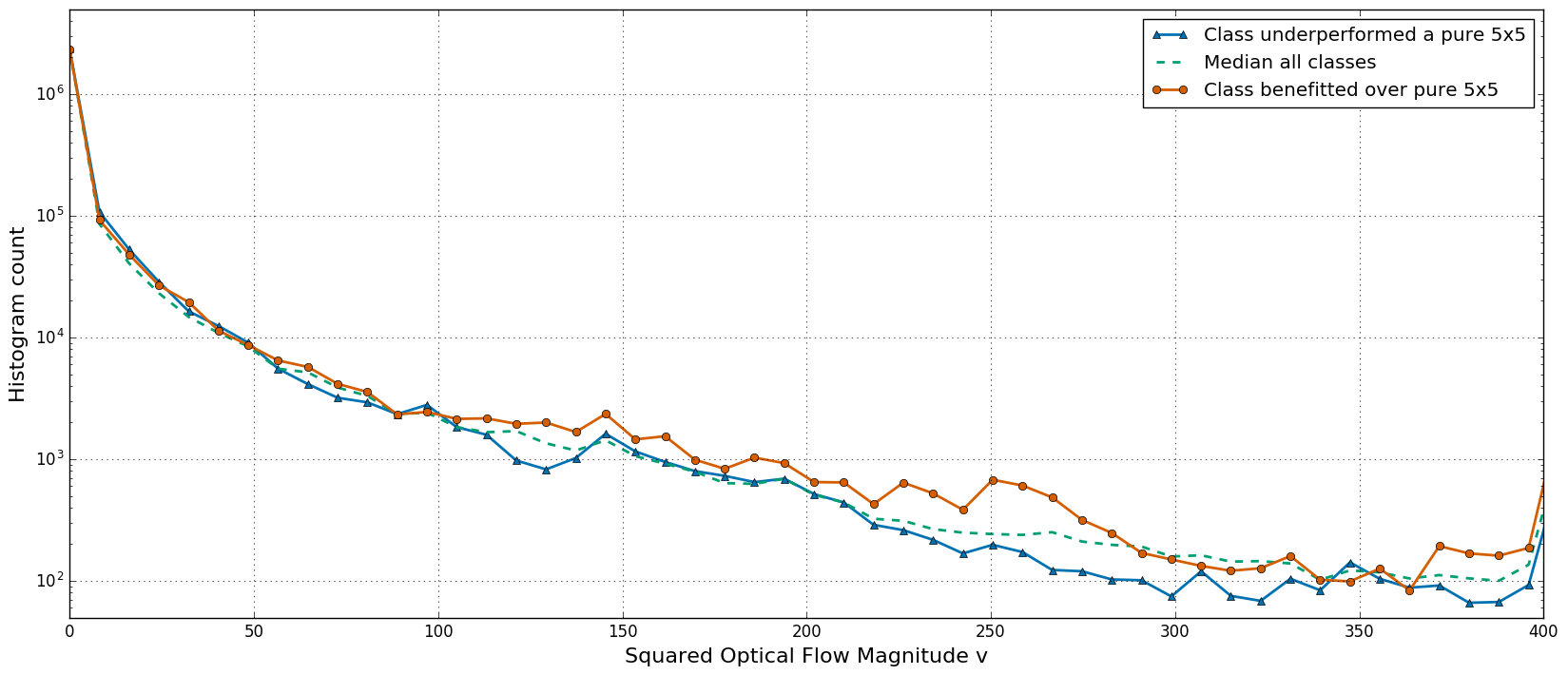}
  \caption{Histogram of squared optical flow velocities $v=\left(x^2 + y^2\right)$ (also see Fig.~\ref{fig:histogram_velo_3}) for a $5\times5$ ConvLSTM baseline and our extension. In comparison to Fig.~\ref{fig:histogram_velo_3}, the red and blue lines diverge earlier, approximately between $100\leq v \leq 300$. This is logical; our system performs better than a pure $5\times5$ ConvLSTM baseline when there are fewer high velocities and more small velocities.}
  \label{fig:histogram_velo_5}
\end{figure*}

To investigate this more closely, we collected the sets of UCF-101 classes in which our system performed significantly worse or better (i.e., classes with a difference of at least 10\% accuracy) and computed the magnitude of the optical flow in these videos. Taking the median over the respective sets of classes, we can investigate the distribution of squared velocities $v=x^2 + y^2$, where $x,y$ are the components of the two-dimensional optical flow. The histogram for comparison with the $3\times3$ ConvLSTM baseline can be found in Fig.~\ref{fig:histogram_velo_3}, and the histogram for comparison with the $5\times5$ ConvLSTM baseline is in Fig.~\ref{fig:histogram_velo_5}.

Our improvements in terms of classification accuracy coincide with a \emph{larger proportion} of \emph{higher} velocities in the optical flow, whereas the failure cases coincide with a reduced amount of these speeds. More specifically, the divergence occurs for $200\leq v \leq 400$ compared with the $3\times3$ baseline and for $100 \leq v\leq 300$ compared with the $5\times5$ baseline. This makes sense; we can improve performance over the $3\times3$ ConvLSTM particularly if more higher velocities than those that a $3\times3$ convolutional kernel can capture are present. Similarly, to improve performance over the $5\times5$ ConvLSTM, more medium velocities must be present because only these can be captured by the smaller convolutional kernel.

\subsubsection{Computational complexity}
We outline the computational impact of our multi-kernel approach in Table~\ref{tbl:complexity_timings}. All timings refer exclusively to the ConvLSTM layer (i.e., they do not include the feature extractor). As expected, the processing time for mixing kernels of two sizes is between the required times for the traditional ConvLSTM layers of the respective kernel sizes. The overhead is minimal and in the order of 1\%. 

We also computed the number of parameters in Table~\ref{tbl:complexity_parameters}, again only for the ConvLSTM layer. The results underline the necessity of our proposed changes, because larger filters require tens of millions of parameters. To clarify, consider that the 25 million parameters of a $7\times7$ ConvLSTM make up approximately 18\% of the total parameters of VGG-16. Mixing kernels of different sizes enables reduction of the parameter count. This count can be further reduced by adding $1\times1$ bottleneck layers ahead of larger convolutional kernels.

\subsubsection{Global motion and local motion} In Section~\ref{sec:motion}, we studied how convolutional kernels of different sizes reacted to moving objects. However, our investigation was limited to global motion, in which an object is uniformly moving in one direction. Here, we investigate if the same is true for local motion using human gestures as examples. 
Therefore, we use an action recognition dataset with pose annotation, JHMDB \cite{lit:jhmdb}. Human poses in JHMDB are represented by 15 two-dimensional pose joint coordinates.

\begin{figure}
  \subfloat[]{\includegraphics[width=.49\columnwidth]{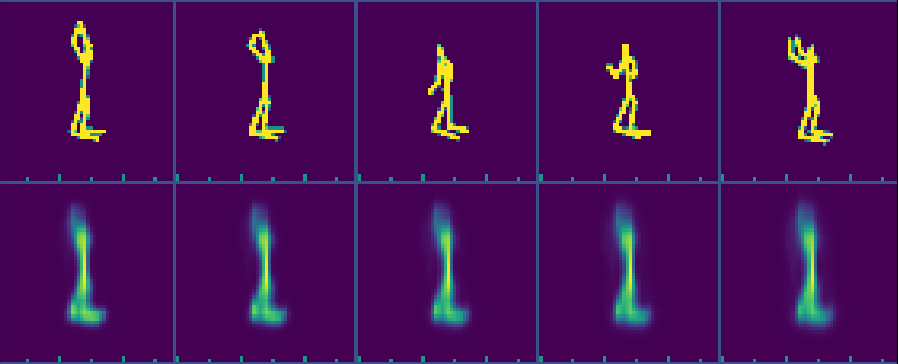}}\hspace{0.01in}
  \subfloat[]{\includegraphics[width=.49\columnwidth]{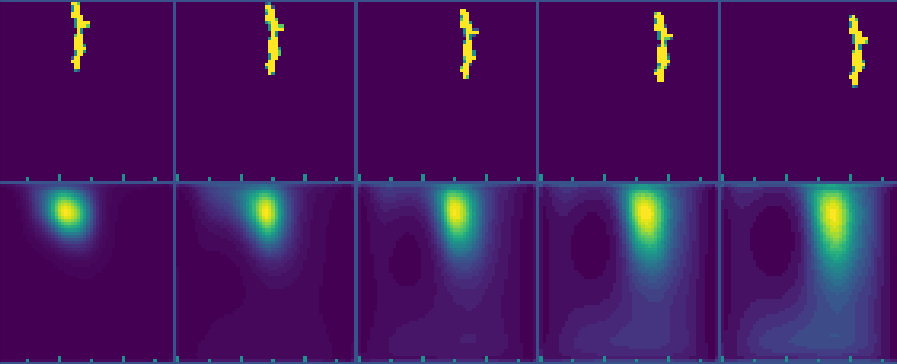}}
  \caption{Visualization of (a) Local motion and (b) global motion prediction on JHMDB. Top row: Groundtruth, bottom row: Prediction.}
  \label{fig:vis_local_global}
\end{figure}

\begin{table}
\centering
  \caption{Prediction loss (SCE) for local motion in human gestures on the JHMDB dataset. Small kernels provide slightly better performance, however the impact is smaller than for global motion. \emph{Normal speed} refers to sampling the dataset at every frame, whereas \emph{fast speed} indicates sampling only every third frame.}
  \begin{tabular}{|l|r|r|}
    \hline
    \textbf{Configuration} &  \textbf{Normal speed} & \textbf{Fast speed} \\ \hline
    $3\times3\times256$ & 1288 & 1521 \\ \hline
    $5\times5\times256$ & 1295 & 1521 \\ \hline
    $7\times7\times256$ & 1331 & 1553 \\ \hline
  \end{tabular}
  \label{tbl:gesture_motion}
\end{table}

Our goal is to analyze local motion. To simplify our experiments, we first eliminated texture information as well as global motion information as follows: we preprocessed the pose data by subtracting the coordinates of the center joint, here ``belly.'' This was followed by centering and scaling. To build an input representation that can be processed by a ConvLSTM, we generated a skeleton for each pose by drawing lines between pose joints on a black background. Fig.~\ref{fig:vis_local_global} shows examples of this dataset (top row), and visualizes prediction results for both local and global motion. 
To test the influence of the kernel size, we again used a prediction task as in Section~\ref{sec:motion}. Images of size $56\times56$ were max-pooled and processed by a two-layer ConvNet first. The resulting $28\times28$ features were fed to a ConvLSTM with $C=256$. The prediction task was implemented in an encoder--decoder fashion, and the used loss was the sigmoid cross entropy. 

Quantitatively, the results in Table~\ref{tbl:gesture_motion} show a slight improvement for smaller kernel sizes. Compared to the global motion experiment (see Fig.~\ref{fig:correlation_result}), the effect is significantly smaller. We argue that gesture motion typically occurs in a more confined area. By providing a mixture of small and large kernels, our system is robust to both global and local motion. 

\vspace{-0.1in}
\subsection{Qualitative results}
\label{sec:qualitative}
Here, we discuss visualizations that provide insight into our proposed extensions. In particular, we discuss saturation problems and our attention scheme.


\begin{figure}
  \centering
    \subfloat[ConvLSTM $3\times3$]{\includegraphics[width=.5\columnwidth]{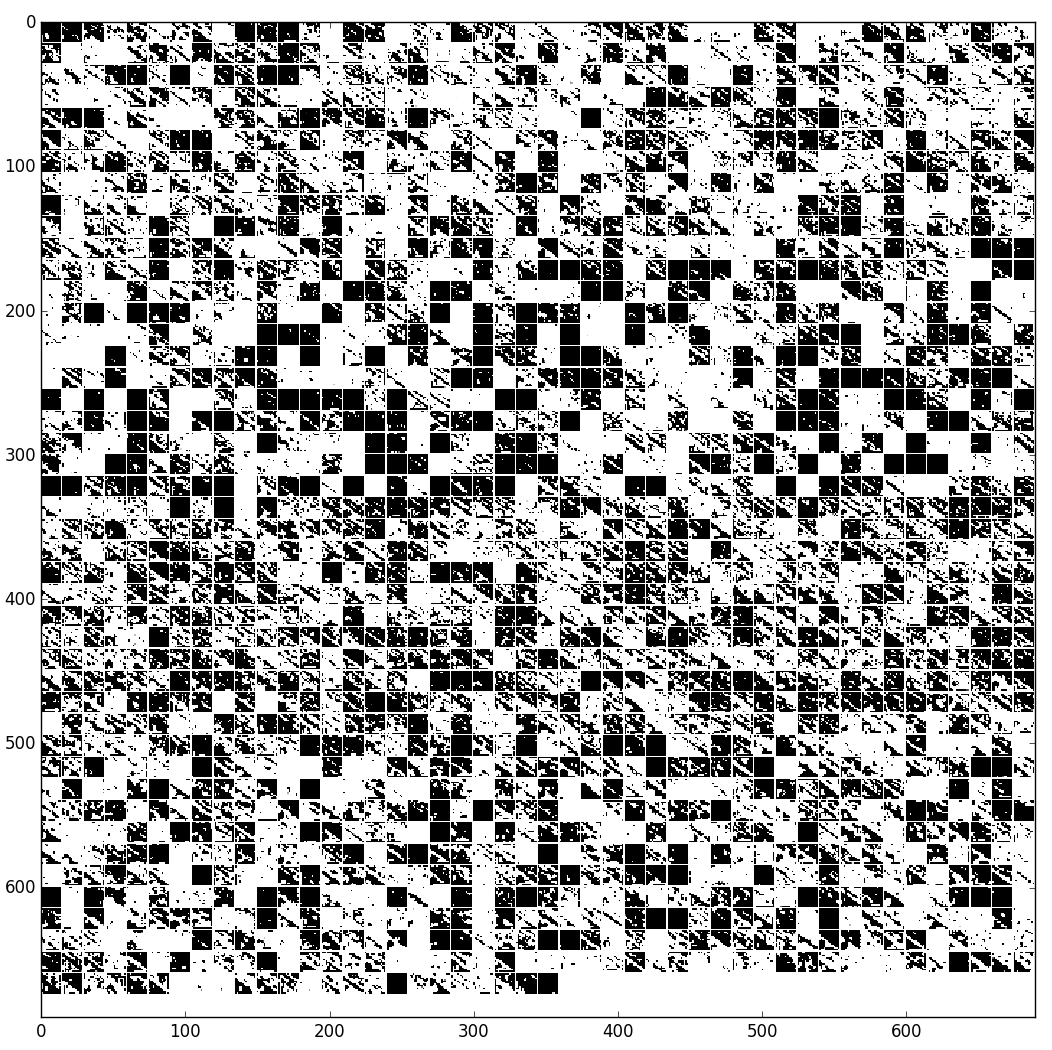}} 
    \subfloat[ConvLSTM $5\times5$]{\includegraphics[width=.5\columnwidth]{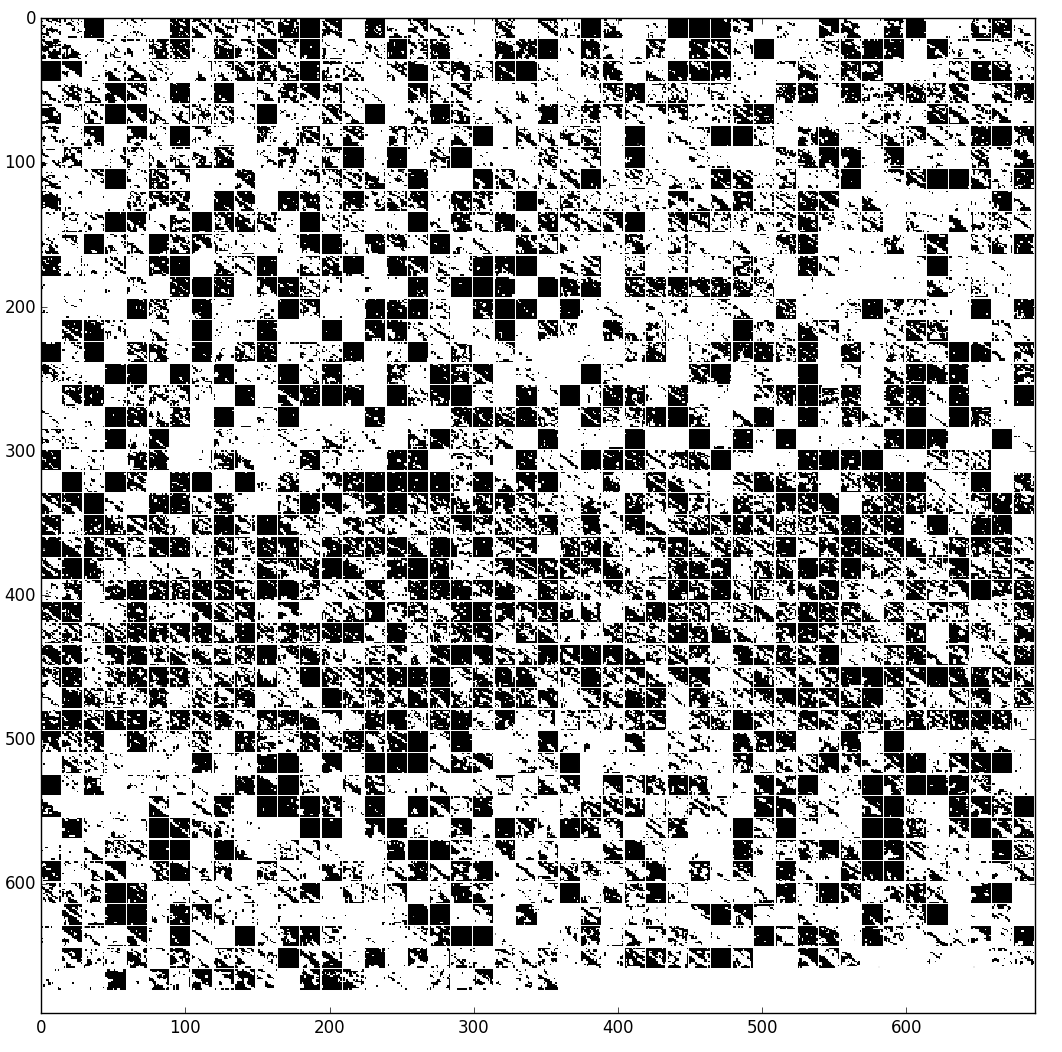}}
    \newline
    \subfloat[Multi-kernel, no additional activation function.]{\includegraphics[width=.5\columnwidth]{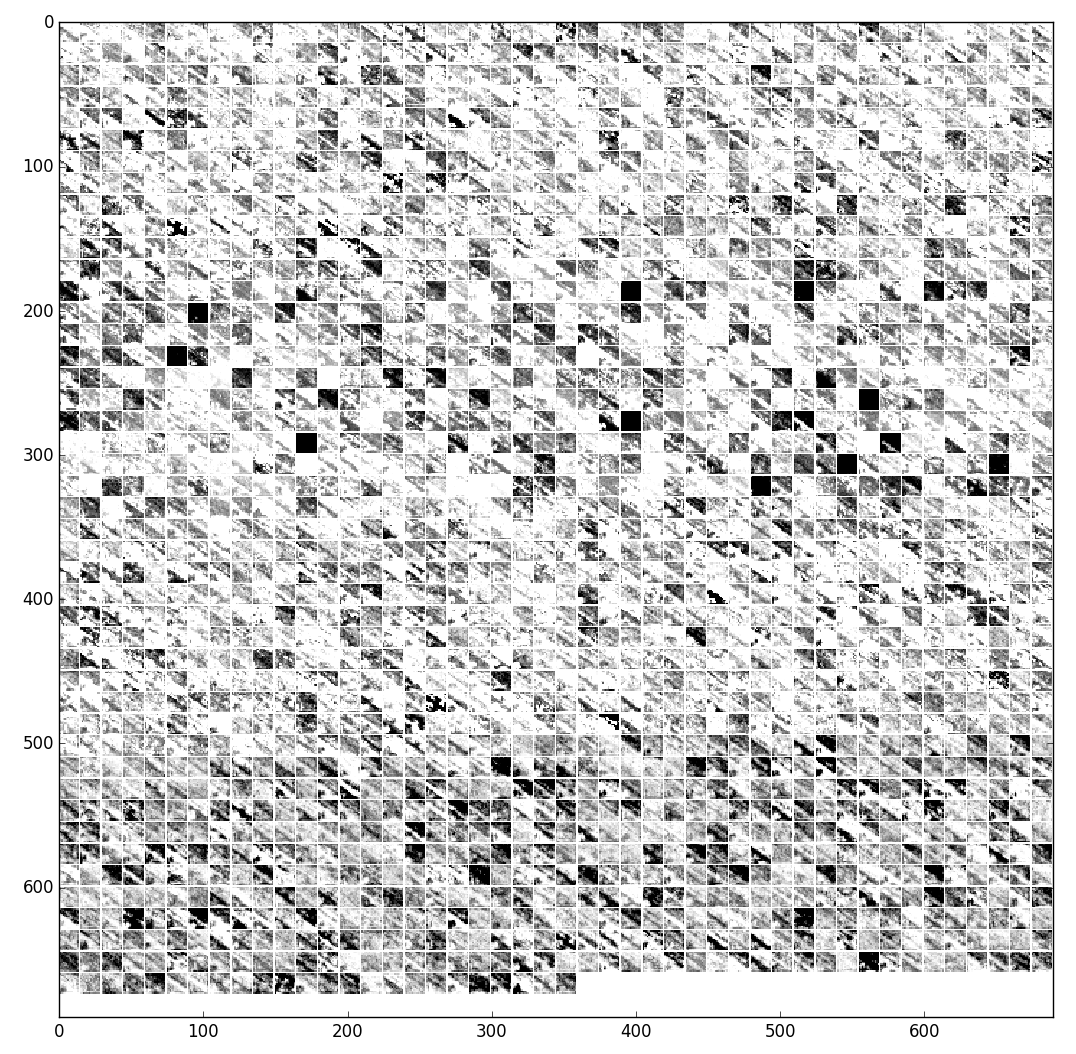}}
    \subfloat[Multi-kernel, $\tanh$ activation.]{\includegraphics[width=.5\columnwidth]{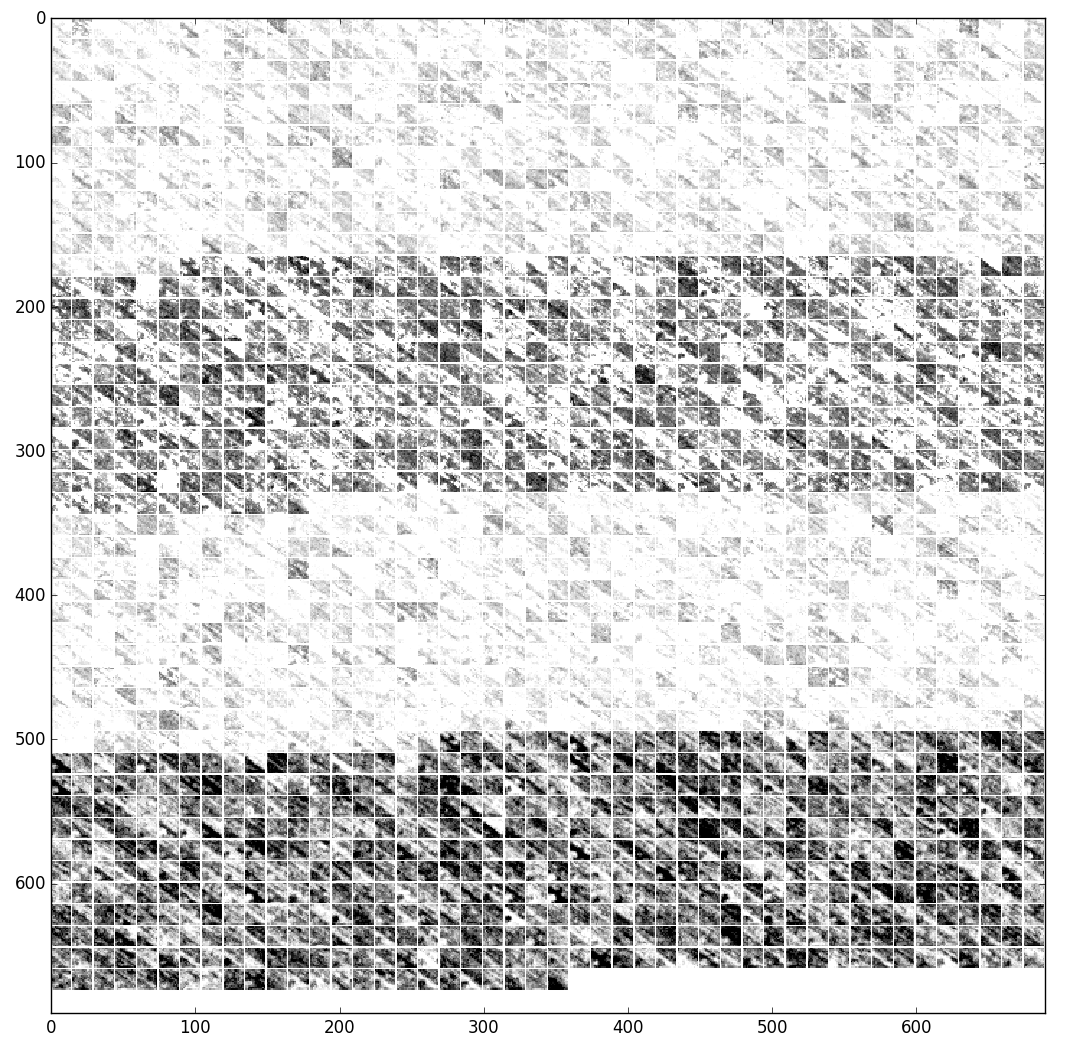}}
    \newline
  \caption{Visualization of the three LSTM gates (\emph{input}, \emph{forget}, \emph{output}) and cell activation (\emph{tanh} term in cell-state) features (in that order). Notably, both baselines (a, b) are oversaturated, taking on only black-and-white color values. Adding the nonlinearity improved homogeneity within gates (c, d). Best viewed digitally.}
  \label{fig:vis_filter}
\end{figure}

\subsubsection{Saturation in ConvLSTM}

Gate activations in our baseline ConvLSTM models typically took on extreme values. Using our multi-kernel scheme with additional \emph{depth}, i.e., the $1\times1$ convolution in our model, this was not the case. We visualize this matter in Fig.~\ref{fig:vis_filter}, where we show the feature maps (after applying $\sigma$ and $\tanh,$ respectively). Fig.~\ref{fig:vis_filter} a) and \ref{fig:vis_filter} b) appear black and white due to the extreme values, whereas our approach in \ref{fig:vis_filter} c) and \ref{fig:vis_filter} d) produces well-balanced features. 

We also note a difference in homogeneity between Fig.~\ref{fig:vis_filter} c) and \ref{fig:vis_filter} d), which differs only in an additional $\tanh$ activation function before the $1\times1$ convolutional layer. These findings clarify which feature map is associated with which gate; \emph{input gate} activations are generally high and encourage saving of input information, whereas \emph{forget gate} activations have a tendency to be lower. Note that cell activations have a value range of (-1,1) and therefore appear darker and are not directly comparable to the control gates.

\subsubsection{Multi-kernel attention}
\label{sec:attention_eval}

\begin{figure}
  \centering
  \subfloat[Input sequence.]{\includegraphics[width=\columnwidth]{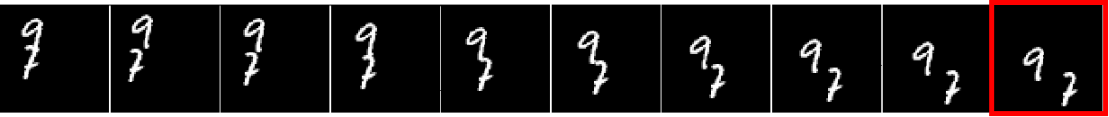}}
  \newline
  \subfloat[Masks for $5\times5$ kernel.]{\includegraphics[width=.5\columnwidth]{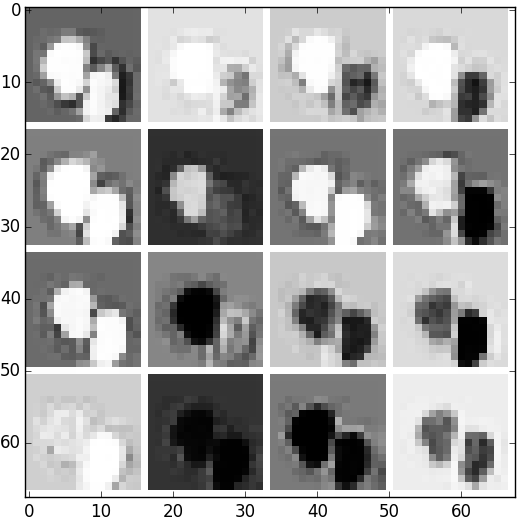}}
  \subfloat[Masks for $7\times7$ kernel.]{\includegraphics[width=.5\columnwidth]{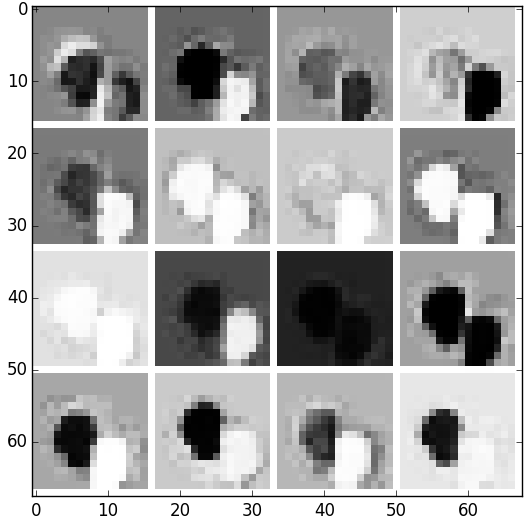}}
  \caption{Visualization of the (16-channel) input attention masks on a modified MovingMNIST dataset for the (a) last frame in sequence. Black areas represent 0 and removed information, whereas white areas represent 1 and admit information. Notice there is a preference in (b) to admit the slow moving digit (``9'') and in (c) to admit the fast moving digit (``7'').}
  \label{fig:attention_mask}
\end{figure}

In Section~\ref{sec:method}, we discussed a flow-based attention mechanism specifically for multiple kernel systems. To visualize the attention masks, we chose the MovingMNIST dataset, which is a synthetic dataset that is based on the handwritten digits of the well-known MNIST dataset. Sequences of 20 frames were generated by picking a $22 \times 22$ digit from MNIST and pasting it into a $64 \times 64$ empty bitmap. A speed and an orientation were assigned to the digit. On contact with the image borders, the digit ``bounced'' back. Several variations of this dataset exist; here, we use the sequence generator supplied\footnote{Website: \url{http://home.cse.ust.hk/~xshiab/}} with the original ConvLSTM work \cite{lit:convlstm} to generate a sequence with two digits, where one digit moves at an average speed of 1 and the other at an average speed of 8.

Following \cite{lit:convlstm} and \cite{lit:unsupervisedlstm}, we trained a future predictor based on our multi-kernel ConvLSTM. As in the original ConvLSTM \cite{lit:convlstm}, we patchified the input. The encoding ConvLSTM was fitted with a $5\times5$ and a $7\times7$ kernel, both using the attention mechanism. 

\subsubsection{Discussion} We visualize the learned masks and the corresponding frame sequence in Fig.~\ref{fig:attention_mask}. Due to the patchification, each mask had $16$ channels. Visual inspection reveals that the slower digit (the ``9'') is more frequently associated with the smaller kernel (i.e., it is represented by a white blob; white represents a high activation value) in Fig.~\ref{fig:attention_mask} b) a total of 10 out of 16 times. The opposite is true for the larger kernel in Fig.~\ref{fig:attention_mask} c), where the fast-moving digit (``7'') is represented 11 out of 16 times by a white blob. This clearly shows the tendency of the larger kernel to associate with faster objects and supports our initial argument on multi-kernel ConvLSTM.

%% file: images/stack.pdf_tex
\begingroup%
  \makeatletter%
  \providecommand\color[2][]{%
    \errmessage{(Inkscape) Color is used for the text in Inkscape, but the package 'color.sty' is not loaded}%
    \renewcommand\color[2][]{}%
  }%
  \providecommand\transparent[1]{%
    \errmessage{(Inkscape) Transparency is used (non-zero) for the text in Inkscape, but the package 'transparent.sty' is not loaded}%
    \renewcommand\transparent[1]{}%
  }%
  \providecommand\rotatebox[2]{#2}%
  \ifx\svgwidth\undefined%
    \setlength{\unitlength}{323.34705743bp}%
    \ifx\svgscale\undefined%
      \relax%
    \else%
      \setlength{\unitlength}{\unitlength * \real{\svgscale}}%
    \fi%
  \else%
    \setlength{\unitlength}{\svgwidth}%
  \fi%
  \global\let\svgwidth\undefined%
  \global\let\svgscale\undefined%
  \makeatother%
  \begin{picture}(1,0.73750256)%
    \put(0,0){\includegraphics[width=\unitlength,page=1]{images/stack.pdf}}%
    \put(-0.00298542,0.075){\color[rgb]{0,0,0}\rotatebox{90}{\makebox(0,0)[lt]{\begin{minipage}{0.13607669\unitlength}\raggedright Frozen\end{minipage}}}}%
    \put(-0.00295784,0.4){\color[rgb]{0,0,0}\rotatebox{90}{\makebox(0,0)[lt]{\begin{minipage}{0.11133547\unitlength}\raggedright Learned\end{minipage}}}}%
    \put(0.73772475,0.705){\color[rgb]{0,0,0}\makebox(0,0)[lt]{\begin{minipage}{0.53193615\unitlength}\raggedright Softmax\end{minipage}}}%
    \put(0.73690066,0.605){\color[rgb]{0,0,0}\makebox(0,0)[lt]{\begin{minipage}{0.38348885\unitlength}\raggedright FC-LSTM\end{minipage}}}%
    \put(0.73667931,0.495){\color[rgb]{0,0,0}\makebox(0,0)[lt]{\begin{minipage}{0.43297129\unitlength}\raggedright Pooling\end{minipage}}}%
    \put(0.7386266,0.38232094){\color[rgb]{0,0,0}\makebox(0,0)[lt]{\begin{minipage}{0.4577125\unitlength}\raggedright ConvLSTM\end{minipage}}}%
    \put(0.73972326,0.18319212){\color[rgb]{0,0,0}\makebox(0,0)[lt]{\begin{minipage}{0.43297129\unitlength}\raggedright VGG-16\\ features\end{minipage}}}%
  \end{picture}%
\endgroup%

%% file: 05.conclusion/conclusion.tex
\section{Conclusion}
\label{sec:conclusion}

We analyzed the problem of different speeds in ConvLSTM and proposed replacing the single convolutional kernel with a set of kernels or a small network. To the best of our knowledge, we are the first to propose such an extension. We also presented an attention-based method that is specifically tailored to our system. Our results were tested using the popular UCF-101 and Sports-1M action recognition datasets, and the findings were discussed using qualitative analysis.

\subsubsection{Future Work}
In Section~\ref{sec:mask}, we described an attention scheme for use with multiple kernels. We limited the investigation to the input-to-hidden operations $\mathbf{W}_{x*}$ and demonstrated the behavior of multiple kernel attention exemplary on MovingMNIST. Future studies may wish to examine if an extension to hidden-to-hidden ($\mathbf{W}_{h*}$) operation is beneficial for multiple kernel systems. An initial idea is to introduce an additional latent state to the ConvLSTM layer, which is manipulated by input and forget gates in a manner similar to the cell-state in equation~(9). However, this will come at the expense of higher memory usage.